\documentclass[10pt,twocolumn,letterpaper]{article}

\usepackage{wacv}
\usepackage{times}
\usepackage{epsfig}

\ifwacvfinal
\usepackage[breaklinks=true,bookmarks=false]{hyperref}
\else
\usepackage[pagebackref=true,breaklinks=true,colorlinks,bookmarks=false]{hyperref}
\fi

\usepackage{url}            
\usepackage{booktabs}       
\usepackage{amsfonts}       
\usepackage{nicefrac}       
\usepackage{microtype}      

\usepackage{graphicx}
\usepackage{subcaption}
\usepackage{amsmath}
\usepackage{amssymb}
\usepackage[nameinlink,noabbrev]{cleveref}

\def\pdata{p_{\rm{d}}}
\def\pfake{p_{\rm{g}}}
\def\punlab{p_{\rm{u}}}

\def\wacvPaperID{249} 

\wacvfinalcopy 
\ifwacvfinal
\def\assignedStartPage{9876} 
\fi

\ifwacvfinal
\else
\fi

\begin{document}

\title{A Multi-Class Hinge Loss for Conditional GANs}

\author{Ilya Kavalerov\\
University of Maryland, College Park\\
{\tt\small ilyak@umd.edu}
\and
Wojciech Czaja\\
University of Maryland, College Park\\
{\tt\small czaja@umd.edu}
\and
Rama Chellappa\\
Johns Hopkins University, Baltimore\\
{\tt\small rchella4@jhu.edu}
}

\maketitle

\begin{abstract}
We propose a new algorithm to incorporate class conditional information into the critic of GANs via a multi-class generalization of the commonly used Hinge loss that is compatible with both supervised and semi-supervised settings. We study the compromise between training a state of the art generator and an accurate classifier simultaneously, and propose a way to use our algorithm to measure the degree to which a generator and critic are class conditional. We show the trade-off between a generator-critic pair respecting class conditioning inputs and generating the highest quality images. With our multi-hinge loss modification we are able to improve Inception Scores and Frechet Inception Distance on the Imagenet dataset.
\end{abstract}

\section{Introduction}
Generative Adversarial Networks (GANs) \cite{goodfellow} are an attractive approach to constructing generative models that mimic a target distribution, and have shown to be capable of learning to generate high-quality and diverse images directly from data \cite{BigGAN}.
Conditional GANs (cGANs) \cite{mirza} are a type of GAN that use conditional information such as class labels to guide the training of the discriminator and the generator.
Most frameworks of cGANs either augment a GAN by injecting (embedded) class information into the architecture of the real/fake discriminator \cite{projdisc}, or by adding an auxiliary loss that is class based \cite{ACGAN}.

We describe an algorithm that uses both a projection discriminator and an auxiliary classifier with a loss that ensures generator updates are always class specific.
Rather than training  with a function that measures the information theoretic distance between the generative distribution and one target distribution, we generalize the successful hinge-loss \cite{geomGAN} that has become an essential ingredient of state of the art GANs \cite{salismans,BigGAN} to the multi-class setting and use it to train a single generator-classifier pair \cite{salismans}.
While the canonical hinge loss made generator updates according to a class agnostic margin learned by a real/fake discriminator \cite{geomGAN}, our multi-class hinge-loss GAN updates the generator according to many classification margins.
With this modification, we are able to accelerate training compared to other GANs with auxiliary classifiers by performing only 1 D-step per G-step, and we improve Inception and Frechet Inception Distance Scores on Imagenet at $128\times128$ on a SAGAN baseline.
\begin{figure*}[t]
\centering
\includegraphics[width=1\linewidth]{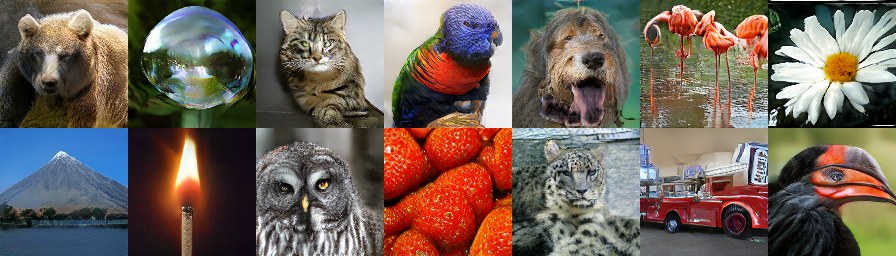}
\caption{ The proposed MHGAN generates images while leveraging the information from the margins of a classifier. All the images in this figure are conditionally sampled from our MHGAN.}
\end{figure*}
\subsection{Background}
A GAN \cite{goodfellow} is a framework to train a generative model that maps random vectors $z\in\mathcal{Z}$ into data example space $x\in \mathcal{X}$ concurrently with a discriminative network that evaluates its success by judging examples from the dataset and generator as real or fake. The GAN was originally formulated as the minimax game:
\begin{multline}
\label{eq:origGAN}
    \max_D  E_{x\sim \pdata} [ \log(D(x)) ] 
         + E_{z\sim p_z} [ \log(1-D(G(z))) ], \\
    \min_G  E_{z\sim p_z } [ \log(1-D(G(z))) ], \ \ \ \ \ \ \ \ \ \ \ \ \ \ \ \ \ \ \ \ \ \ \ \ \ \ \ \ \ \ 
\end{multline}
\noindent
where $\pdata$ is the real data distribution consisting of examples $x\in\mathcal{X}$, $p_z$ is the latent distribution over the latent space $\mathcal{Z}$, $G: \mathcal{Z}\to\mathcal{X}$ is the generator neural network, and $D: \mathcal{X}\to [0,1]$ is the discriminator neural network.
The GAN model transfers the success of the deep discriminative model $D$ to the generative model $G$ and succeeds in generating impressive samples $G(z)$. 
\subsection{Loss functions for training GANs}
We rewrite the GAN optimization problem in a more generic form with minimums \cite{understanding_loss} that is amenable to discussion and programming:
\begin{equation}
\begin{split}
    \min_D & E_{x\sim \pdata} [ f(D(x)) ] 
         + E_{z\sim p_z} [ g(D(G(z))) ], \\
    \min_G & E_{z\sim p_z } [ h(D(G(z))) ].
\end{split}
\end{equation}
\noindent

To regain the minimax objective in \cref{eq:origGAN} we set $f(w)=\log(1+e^{-w})$ and $h(w)= - g(w) =-w-\log(1+e^{-w})$ \cite{understanding_loss}.
This choice minimizes the Jensen-Shannon divergence between $\pdata$ and $\pfake$ \cite{goodfellow}, which denotes the model distribution implicitly defined by $G(z),z\sim p_z$.
The minimax objective however was difficult to train \cite{principled_methods}, and the study of various ways to measure the divergence or distance between $\pdata$ and $\pfake$ has been a source of improved loss functions that make training more stable and samples $G(z)$ of higher quality.

\begin{figure*}[t]
\centering
\begin{subfigure}{.33\textwidth}
  \centering
  \includegraphics[height=3.5cm]{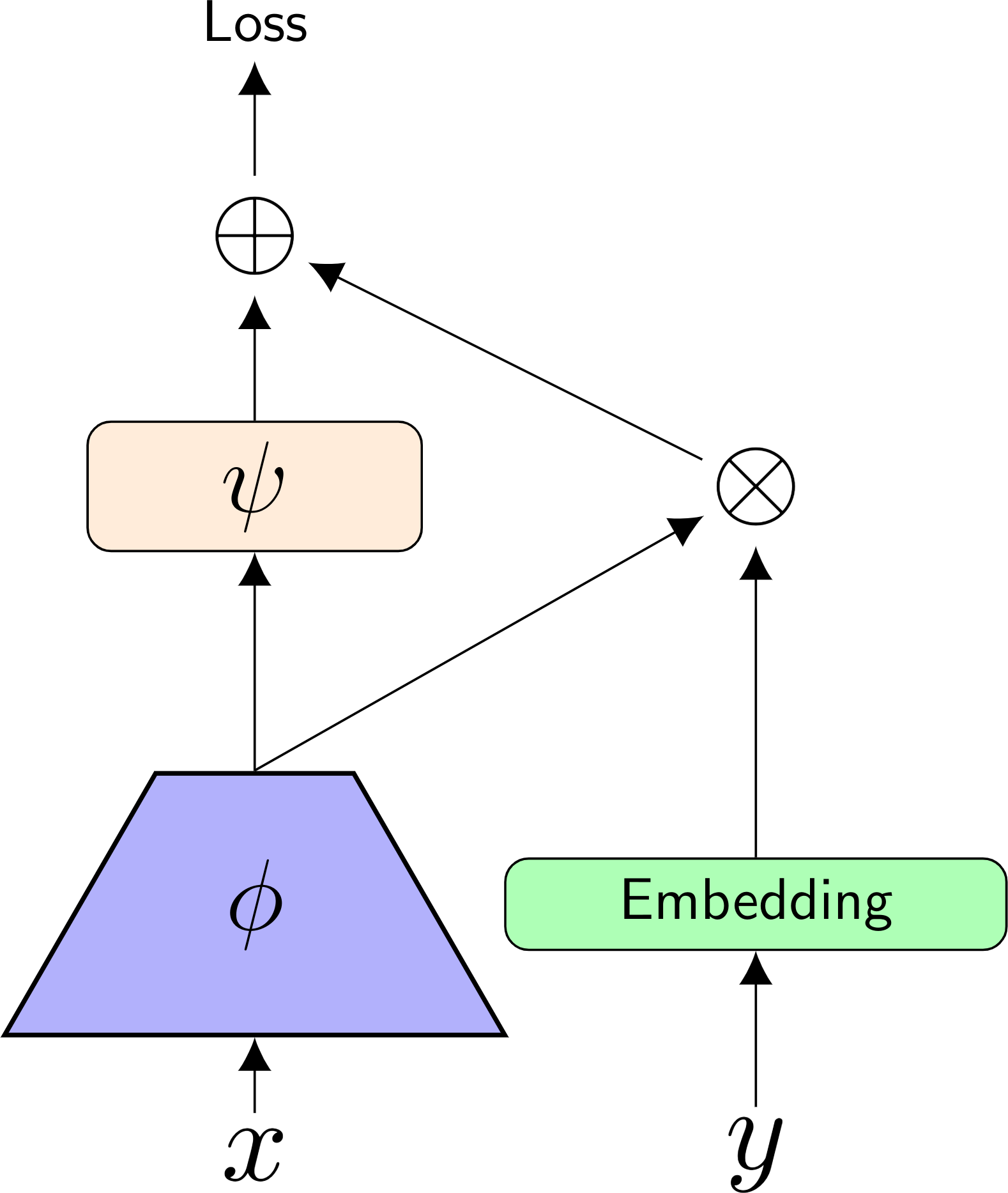}
  \caption{}
  \label{fig:ProjDiscarch}
\end{subfigure}%
\hfill
\begin{subfigure}{.33\textwidth}
  \centering
  \includegraphics[height=3.5cm]{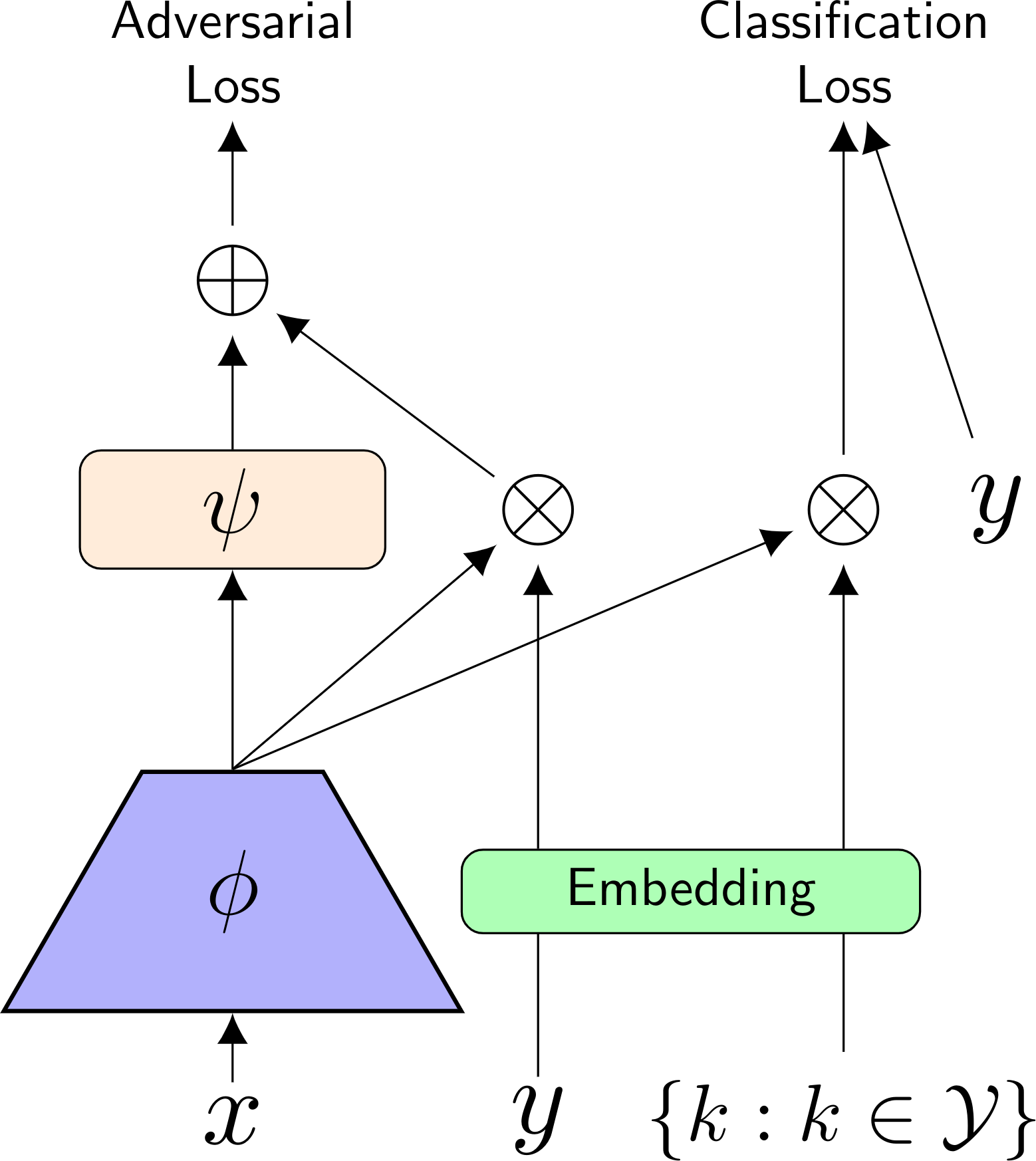}
  \caption{}
  \label{fig:MHSharedGANarch}
\end{subfigure}%
\hfill
\begin{subfigure}{.33\textwidth}
  \centering
  \includegraphics[height=3.5cm]{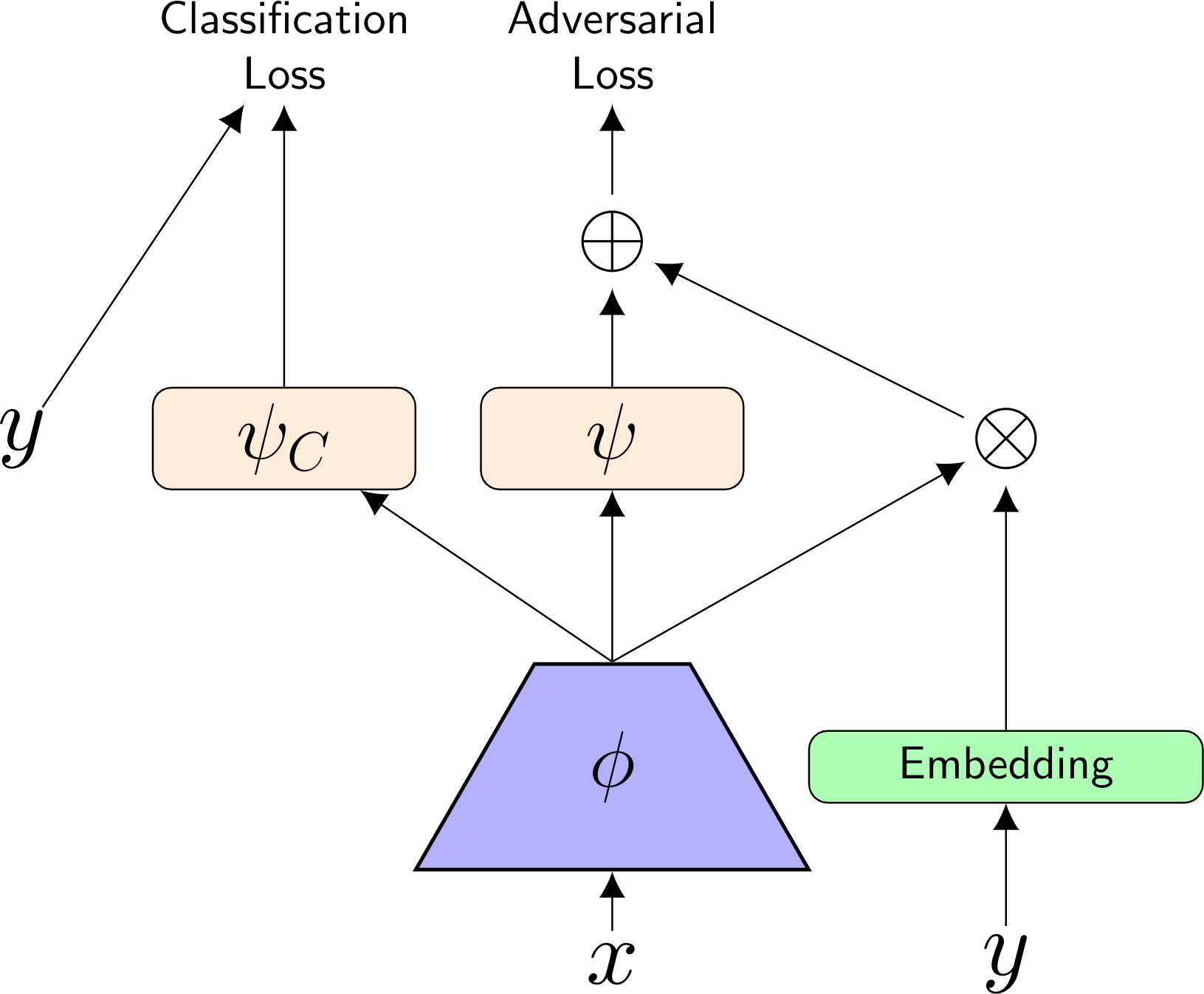}
  \caption{}
  \label{fig:ACGANarch}
\end{subfigure}
   \caption{
   \ref{fig:ProjDiscarch} shows the projection discriminator architecture \cite{projdisc} of our SAGAN. \ref{fig:MHSharedGANarch} shows how the projection discriminator could simultaneously be trained with an auxiliary classifier loss: we train MHSharedGAN this way. \ref{fig:ACGANarch} Is an ACGAN \cite{ACGAN} architecture that also contains a projection discriminator: we train our ACGAN and MHGAN with this architecture.  }
\label{fig:stlerr}
\label{fig:onecol}
\end{figure*}

WGAN set $f(w)= -w$, $h(w) = - g(w) =-w$ and clipped the weights of $D$ to greatly improve the ease and quality of training and reduced the mode dropping problem of GANs; it has the interpretation of minimizing the Wasserstein-1 distance between $\pdata$ and $\pfake$.
Minimizing Wassertstein-1 distance was shown to be a special instance of minimizing the integral probability metric (IPM) between $\pdata$ and $\pfake$ \cite{IPM},
and inspired a mean and covariance feature matching IPM loss (McGAN)\cite{IPM} following the empirical successes of the Maximum Mean Discrepancy objective \cite{MMD_Li2017} and feature matching \cite{salismans}.

The mean feature matching of McGAN has a geometric interpretation: the gradient updates of feature matching for the generator are normal to the separatating hyperplane learned by the discriminator \cite{geomGAN}.
The SVM like hinge-loss choice of $f(w)=\max(0,1-w),\ g(w)=\max(0,1+w)$ and $h(w)=-w$ \cite{geomGAN,Tran2017} has gradients similar to those of McGAN.
When combined with spectral normalization of weights in $D$ \cite{Miyato2018}, the hinge loss greatly improves performance, and has become a mainstay in recent state of the art GANs \cite{BigGAN,SAGAN,projdisc}.
In this work we generalize this hinge loss to a multi-class setting.
\subsection{Supervised training for conditional GANs}
Conditional GANs (cGANs) are a type of GAN that use conditional information \cite{mirza} in the discriminator and generator.
$G$ and $D$ become functions of the pairs $(z\sim p_z,y\sim\pdata)$ and $(x,y)\sim\pdata$, where $y$ is the conditional data, for example the class labels of an image.
In a cGAN with a hinge loss, the discriminator would minimize $L_D$ in \cref{eq:orighinge}, and the generator would minimize $L_G$ in \cref{eq:orighinge} \cite{SAGAN,BigGAN}.
\begin{equation}
\label{eq:orighinge}
\begin{split}
    L_D = & E_{(x,y)\sim \pdata} [ \max(0,1-D(x,y)) ]  \\
        + & E_{z\sim p_z,y\sim \pdata} [ \max(0,1+D(G(z,y),y)) ] \\
        = & L_{D\rm{real}} + L_{D\rm{fake}}, \\
    L_G = & -E_{z\sim p_z,y\sim \pdata } [ D(G(z,y),y) ] .
\end{split}
\end{equation}
We briefly review some work on using conditional information to train the discriminator of GANs, as well as uses of classifiers.

A projection discriminator \cite{projdisc} is a type of conditional discriminator that adds the inner product between an intermediate feature and a class embedding to its final output, and proves highly effective when combined with spectral normalization in $G$ \cite{projdisc,SAGAN,BigGAN}.
Several GANs have used a classifier in addition to, or in place of, a discriminator to improve training.
CatGAN \cite{catgan} replaces the discriminator with a $K$-class classifier trained with cross entropy loss that the generator tries to confuse.
ACGAN \cite{ACGAN} uses an auxiliary classification network or extra classification layer appended to the discriminator, and adds the cross entropy loss from this network to the minimax GAN loss.
Triple GAN \cite{tripleGAN} trains a classifier in addition to a discriminator and updates it with a special minimax type loss.

Improved GAN \cite{salismans} originally proposed using a $K+1$ classifier for semi-supervised learning (SSL) with feature matching loss, and others \cite{odena_ssl} have a used a similar approach for SSL GANs as well.
The single conditional critic architecture of $D: (x,y)\to\mathbb{R}$ is swapped for the classifier architecture $C: x \to \mathbb{R}^{K+1}$, where there are $K$ class labels and an extra label for fake images (the "$+1$") \cite{salismans}. The Improved GAN trains this classifier architecture in a semi-supervised setting with log-likelihood loss, and trains the generator with a class agnostic mean feature matching loss.
BadGAN \cite{badgan} used the Improved GAN to achieve state of the art performance on semi-supervised learning classification and found the aim of having a low classification error on the $K$ real classes is orthogonal to generating realistic examples.
MarginGAN \cite{margingan} used a Triple GAN to train a high-quality classifier with a "bad" GAN \cite{badgan} by decreasing the margins of error of the cross entropy loss for generated images.
There are few works which focus on improving the generator quality in a label limited setting. One approach has used two discriminators, one specializing on labeled and the other on unlabeled data \cite{sricharan2017semi}, another has found that pre-labelling the unlabelled training data with a SSL classifier before GAN training to be a high performance method \cite{lucic_ssl}.

Our proposed multi-hinge GAN (MHGAN) uses a classifier like ACGAN \cite{ACGAN} but instead of using a probabilistic cross entropy loss with the WGAN, we use a multi-hinge loss similar to that in the discriminator.
We demonstrate that adding this loss to the current state of the art SAGAN architecture with projection discrimination improves image quality and diversity, and trains stably at only 1 discriminator step per generator step for both supervised and semi-supervised settings.
We compare our class specific modification with cross entropy loss \cite{ACGAN} with projection discrimination, and with projection discrimination alone. 
Unlike cross-entropy, which has a basis in probabilistic quantities that are not present in WGANs, our multi-hinge loss is completely compatible with the WGAN formulation.
We also find that the task of classification and discrimination should not share parameters, and show how to trade-off image diversity for quality in a shared parameter version MHSharedGAN.
%



\section{Multi-hinge loss}

We propose a multi-hinge loss that can be easily plugged into the popular and state of the art projection discriminator cGAN architecture \cite{projdisc}.
Our fully supervised formulation, motivated in \Cref{sec:mot}, uses the auxiliary classifier setup seen in \Cref{fig:ACGANarch}, but instead of using cross entropy as ACGAN \cite{ACGAN} does to train this classifier we generalize the binary hinge loss \cite{geomGAN,Tran2017} to a multi-class hinge loss developed for SVMs \cite{crammer_singer}, and we use it to train a spectrally normalized WGAN \cite{wgan,wgan2,Miyato2018,SAGAN}. Unlike other ACGANs, our multi-hinge loss formulation only requires 1 Critic step per 1 Generator step greatly speeding up training, and uses a single classifier for all classes in the dataset.

We denote the classifier function as $C: \mathcal{X} \to \mathcal{Y}$ and let $C_k(x)$ denote the $k$th element of the vector output of $C$ for an example $x$, which represents the affinity of class $k$ for $x$.
The Crammer-Singer multi-hinge loss that we propose as an auxiliary term is:
\begin{equation}
\begin{split}
    & L_{D\rm{aux}} = E_{(x,y)\sim \pdata} [ \max(0,1-C_y(x) + C_{\neg y}(x)) ] \\
    & L_{G\rm{aux}} = \\
    &\ \ E_{ \substack{\\ z\sim p_z \\ y\sim \pdata} } [ \max(0,1-C_y( G(z,y) ) + C_{\neg y}( G(z,y) )) ]
\end{split}
\label{eq:mhsimplest}
\end{equation}
\noindent
where $C_{\neg y}(x)$ is the classifier's highest affinity for any label that is "not $y$": $C_{\neg y}(x) = \max_{k\neq y} C_k(x), y=0,1,\dots, K$. We then train with the modified loss:
\begin{equation}
\label{eq:overallhinge}
\begin{split}
    L_{\rm{MH},D} = & L_{D\rm{real}} + L_{D\rm{fake}} + L_{D\rm{aux}}, \\
    L_{\rm{MH},G} = & L_G + \lambda L_{G\rm{aux}} .
\end{split}
\end{equation}
The advantage of a conditional WGAN trained with the auxiliary terms in \cref{eq:mhsimplest} is the main result of this work. In the following sections we discuss the motivation and advantage of this training procedure.
We also provide an SSL formulation in \Cref{sec:ssl}.
\begin{figure}[t]
\centering
\includegraphics[width=0.8\linewidth]{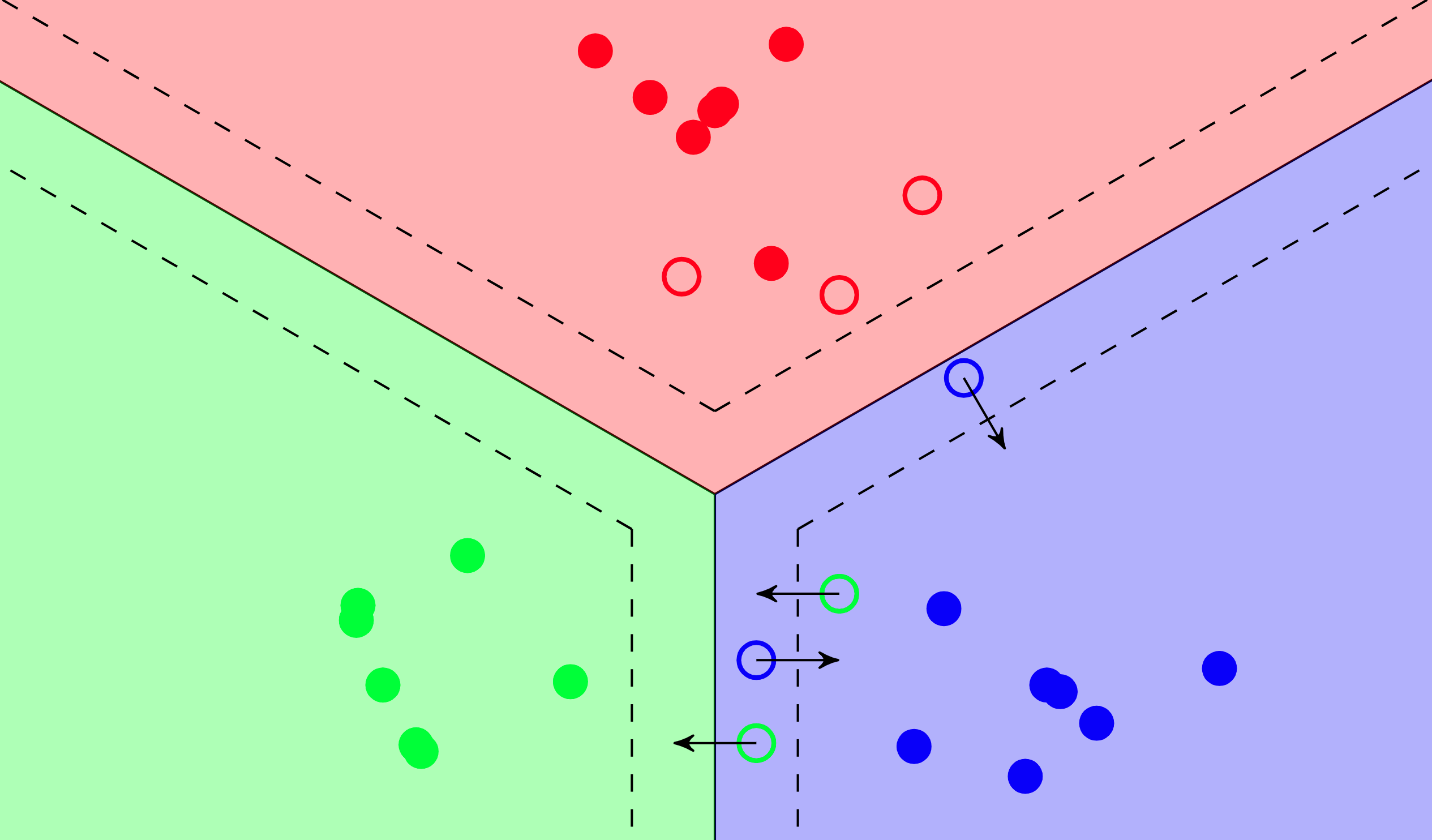}
\caption{This illustration of MHingeGAN training for $L_{G_\mathrm{aux}}$ shows a classifier (solid lines and shaded regions) learned on real samples (filled dots). The class margins (dotted lines) are enforced for generated samples (unfilled dots, colored by their conditioning) by updating the generator with respect to class margins. Green samples are classified incorrectly and the generator update gradient is shown with pointed arrows. Blue generated samples are classified correctly but within the margin, and the gradient is shown. Red generated samples are classified correctly outside of the margin and the MHinge loss has no gradient for these samples.}
\label{fig:intuition}
\end{figure}
\subsection{Motivation \& Intuition}
\label{sec:mot}
A class conditional discriminator should obviously not output "real" when it is conditioned on the wrong class. That is for a pair $(x,y)\sim \pdata$ we expect if our discriminator loss is minimized then so is the quantity: $1-D(x,y)+D(x,k), k\neq y$. 
This quantity is positive for all $k$ so long as the output of the discriminator conditioned on the correct label is larger by at least one than the discriminator conditioned on the rest of the labels.
To explicitly enforce this margin, we could minimize the expectation:
\begin{equation}
    E_{(x,y)\sim \pdata} [ \max(0,1-D(x,y) + \max_{k\neq y}D(x,k) ) ] 
\label{eq:mot1}
\end{equation}
\noindent
This form of a hinge-loss has been used by \cite{crammer_singer} in their formulation of efficient multi-class kernel SVMs.
The ReLU $\max(0,\cdot)$ leads \cref{eq:mot1} to ignore cases where the correct decision is made with a margin more than 1.
We design \cref{eq:mhsimplest} to enforce the margins in \cref{eq:mot1}.
\Cref{fig:intuition} illustrates the effect this loss has on generator training.

Note that a projection discriminator cGAN implicitly has a classifier in it. The output of $D(x,k)$ projects the penultimate features onto an embedding for class $k$. Similarly the vector output of a typical classifier is a matrix multiply of the penultimate features with a features $\times$ class matrix.
A projection discriminator $D(x,k)$ could be turned into a classifier $C(x)$ by using the entire matrix of class embeddings to output an affinity for every class, as shown in \Cref{fig:MHSharedGANarch}.
Creating a classifier this way doesn't increase the parameter count at all, and only increases computation in one layer of $D$ by a constant factor (the number of classes) which we find is completely negligible for the large models typically trained for $64\times64$ images or greater.
However, we find that this sharing of parameters between the classification task in \cref{eq:mot1} and the discrimination task around which the adversarial training centers is disadvantageous.
As we will discuss, it is instead preferable to add an auxiliary classifier via an extra final fully connected layer, the additional memory cost of a features $\times$ class matrix proves to be negligible in our experiments.
\begin{figure*}[t]
\centering

\begin{subfigure}{.33\textwidth}
  \centering
  \includegraphics[width=.98\linewidth]{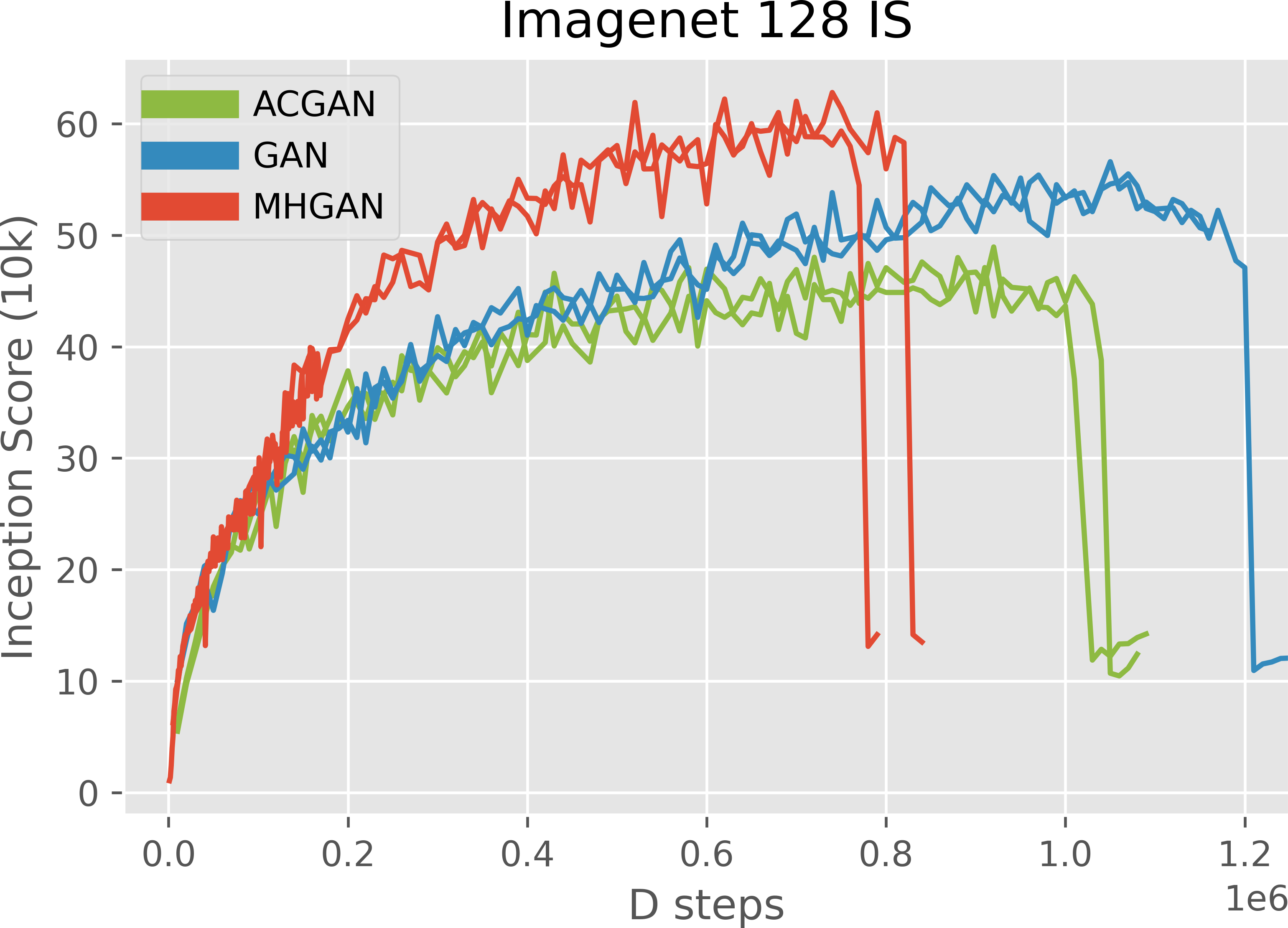}
  \caption{}
  \label{fig:I128_IS}
\end{subfigure}%
\begin{subfigure}{.33\textwidth}
  \centering
  \includegraphics[width=.98\linewidth]{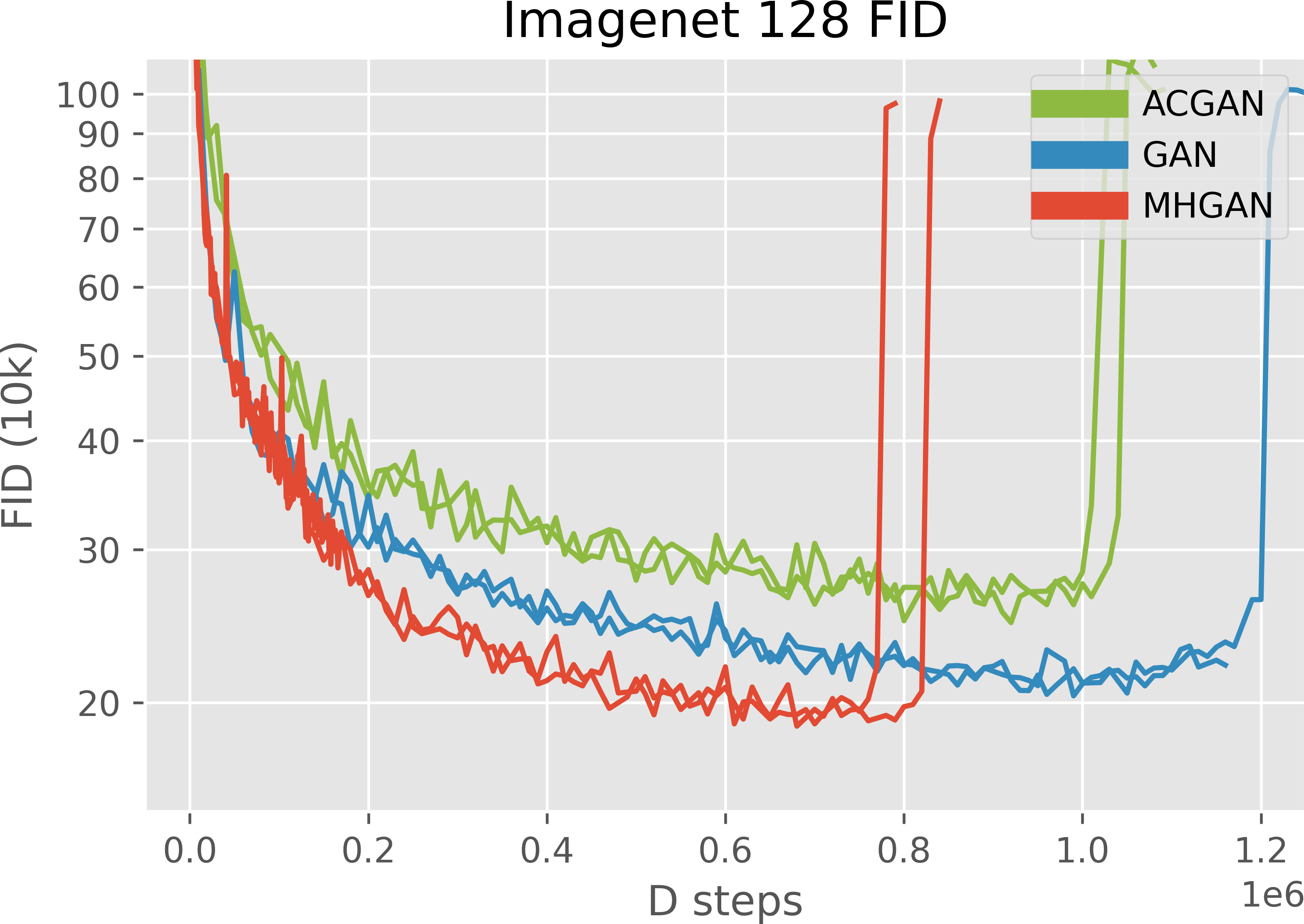}
  \caption{}
  \label{fig:I128_FID}
\end{subfigure}%
\begin{subfigure}{.33\textwidth}
  \centering
  \includegraphics[width=.98\linewidth]{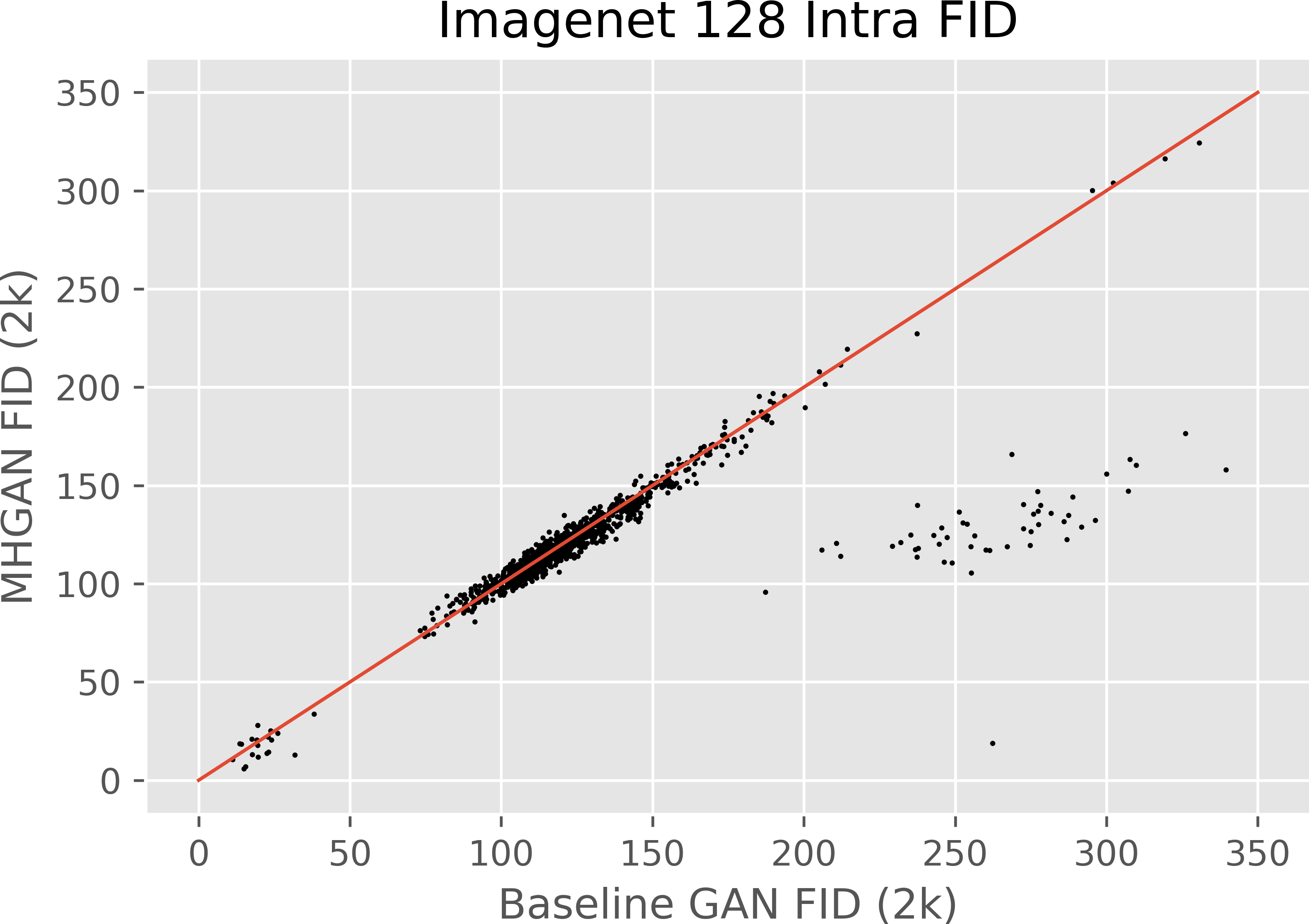}
  \caption{}
  \label{fig:I128_IFID}
\end{subfigure}
\hfill
   \caption{ Evaluation metrics for Imagenet-128. Inception Score and Frechet Inception Distance during training, and a comparison of Intra-class Frechet Inception distance for our best MHGAN model and the best baseline SAGAN we trained. In \ref{fig:I128_IFID} there are 622 classes whose intra-fid improve (each point below the red line is a class for which FID improves) and there are 46 classes in the lower right cluster. }
\label{fig:perf_charts}
\end{figure*}
\subsection{Semi-supervised learning}
\label{sec:ssl}
When additional unlabeled data $x\sim\punlab$ is available, we find that learning with projection discrimination is not stable, that is bypassing the projection discriminator when a label is not available does not lead to successful training. For semi-supervised settings we modify the training procedure in \cref{eq:overallhinge} by using pseudo-labels \cite{lucic_ssl}.
For the discriminator we add the term:
\begin{equation}
L_{D\rm{unlab}} = E_{ x\sim \punlab } [ \max(0,1-D( x,\widetilde{y}_C(x)) ) ]
\label{eq:mhlossDunlab}
\end{equation}
\noindent
where $\widetilde{y}_C(x) = \arg\max_{k\in\mathcal{Y}} C_k(x)$, that is we depend on the classifier that we co-train with the discriminator.
The loss for the generator is left unchanged. Overall for the semi-supervised setting we train with the losses:
\begin{equation}
\label{eq:overallhingessl}
\begin{split}
    L_{\rm{MH,SSL},D} = & \frac{L_{D\rm{real}} + L_{D\rm{unlab}}}{2} + L_{D\rm{fake}} + L_{D\rm{aux}}, \\
    L_{\rm{MH,SSL},G} = & L_{\rm{MH},G} .
\end{split}
\end{equation}
\noindent
A similar loss has previously been used with a WGAN co-trained with a cross entropy auxiliary classifier \cite{lucic_ssl}
\begin{equation}
\label{eq:acssl}
\begin{split}
    L_{\rm{AC,SSL},D} = & \frac{L_{D\rm{real}} + L_{D\rm{unlab}}}{2} + L_{D\rm{fake}}  \\
     & + E_{(x,y)\sim \pdata} [ \log C_y(x)  ], \\
    L_{\rm{AC,SSL},G} = & L_G + \lambda E_{ \substack{\\ z\sim p_z \\ y\sim \pdata} } [ \log C_y( G(z,y) ) ]  .
\end{split}
\end{equation}
\noindent
However we find that the consistency of hinge functions throughout the loss terms leads to more successful training.
%
%
\section{Experiments}
\label{sec:exp}
As our baseline, we use a spectrally normalized \cite{Miyato2018} SAGAN \cite{SAGAN} architecture.
We use this baseline from the publicly available tfgan implementation \cite{tfgan}, and execute experiments on single v2-8 and v3-8 TPUs available on Google TFRC.
This baseline was chosen for its exceptional performance \cite{tfgan}.
On top of this baseline we implement a second baseline, ACGAN, and our MHGAN (we found ACGAN trained without projection discrimination to not be competitive).
The only architectural changes to SAGAN for both of these networks is that a single dense classification layer is added to the penultimate features.
For these networks conditional information is given to $G$ using class conditional BatchNorm \cite{Dumoulin2017,devries} and to $D$ with projection discrimination \cite{projdisc}.
A spectral norm is applied to both $D$ and $G$ during training \cite{SAGAN,projdisc}.
We train our SAGAN baseline with the hinge loss \cite{geomGAN,Tran2017} in \cref{eq:orighinge}.
We train our proposed MHGAN with \cref{eq:overallhinge}.
To better see the advantage of our multi-hinge loss formulation we train an ACGAN baseline with the loss:
\begin{equation}
\label{eq:acgan}
\begin{split}
    L_{\rm{AC},D} = & L_{D\rm{real}} + L_{D\rm{fake}} + E_{(x,y)\sim \pdata} [ \log C_y(x)  ], \\
    L_{\rm{AC},G} = & L_G + \lambda  E_{ \substack{\\ z\sim p_z \\ y\sim \pdata} } [ \log C_y( G(z,y) ) ] .
\end{split}
\end{equation}
\noindent
We also evaluate our multi-hinge formulation for semi-supervised settings and similarly train a MHGAN-SSL model with \cref{eq:overallhingessl}, and compare it with an ACGAN-SSL model trained with \cref{eq:acssl} (SAGAN was not able to train stably without an auxiliary classifier).
\begin{figure*}[t]
\centering
\begin{subfigure}{.2\textwidth}
  \centering
  \includegraphics[width=.98\linewidth]{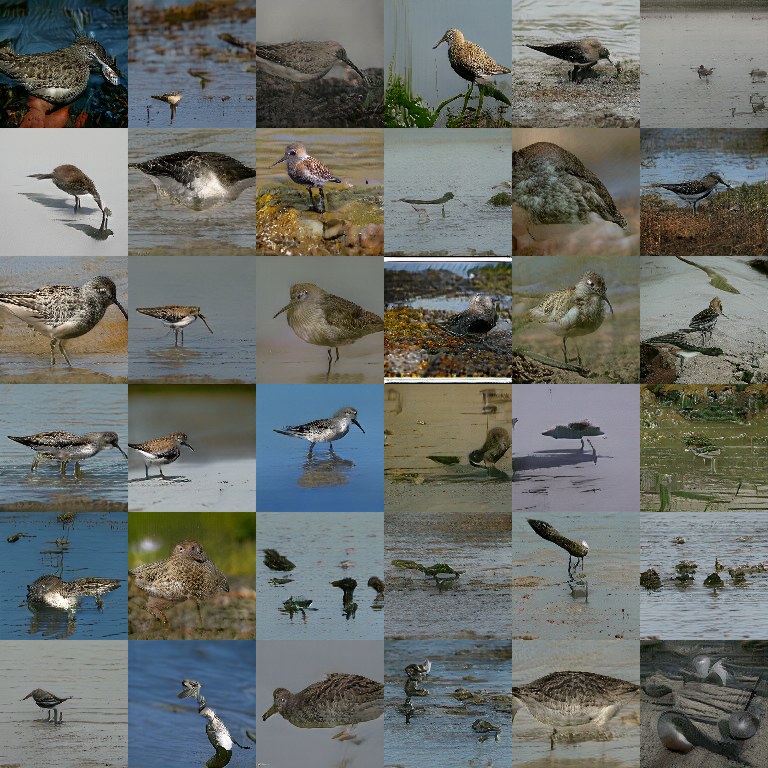}
  \caption{MHGAN IFID 134.82}
\end{subfigure}%
\begin{subfigure}{.2\textwidth}
  \centering
  \includegraphics[width=.98\linewidth]{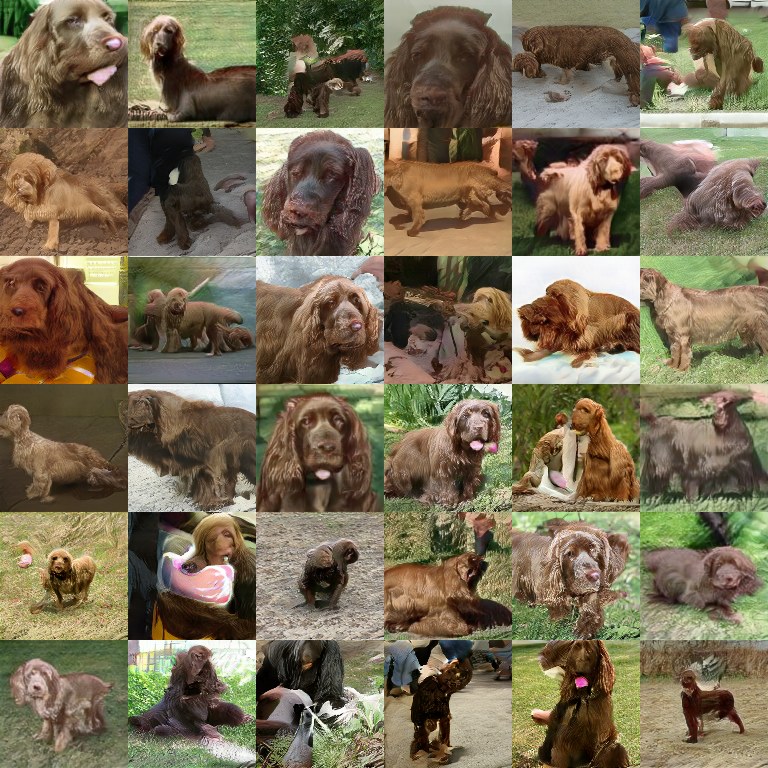}
  \caption{MHGAN IFID 126.52}
\end{subfigure}%
\begin{subfigure}{.2\textwidth}
  \centering
  \includegraphics[width=.98\linewidth]{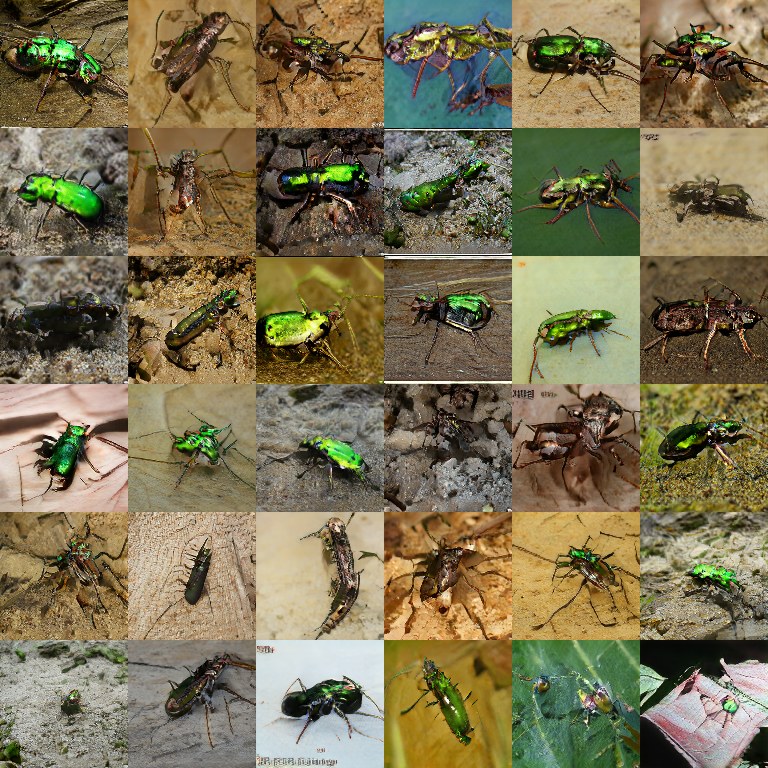}
  \caption{MHGAN IFID 158.00}
\end{subfigure}%
\begin{subfigure}{.2\textwidth}
  \centering
  \includegraphics[width=.98\linewidth]{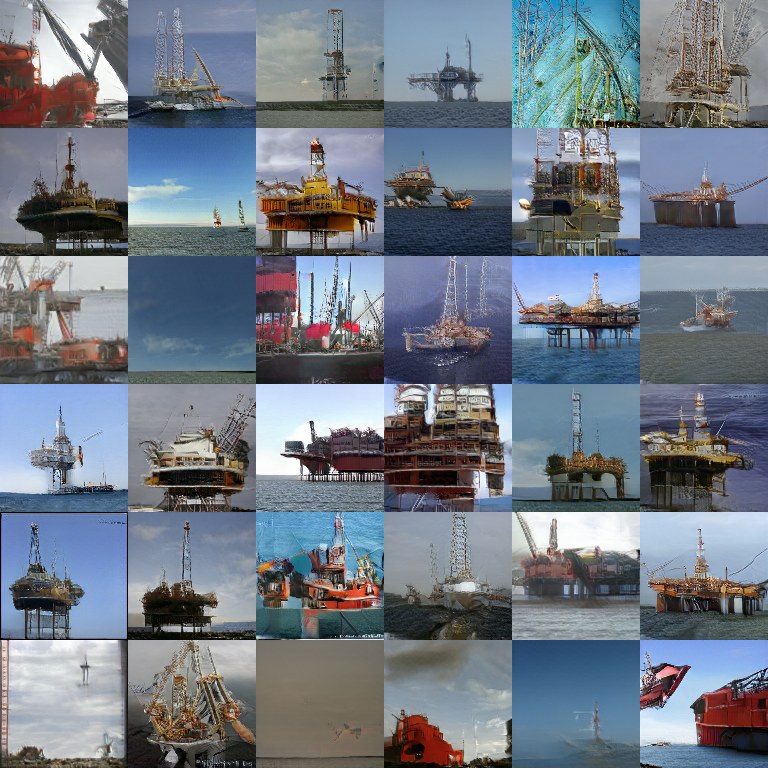}
  \caption{MHGAN IFID 147.14}
\end{subfigure}%
\begin{subfigure}{.2\textwidth}
  \centering
  \includegraphics[width=.98\linewidth]{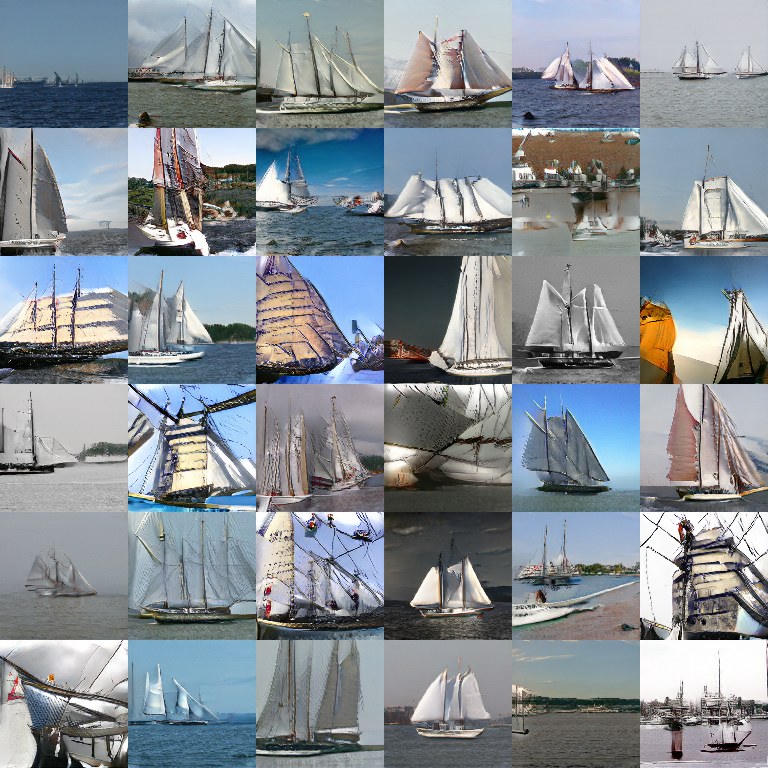}
  \caption{MHGAN 132.38}
\end{subfigure}
\hfill
\begin{subfigure}{.2\textwidth}
  \centering
  \includegraphics[width=.98\linewidth]{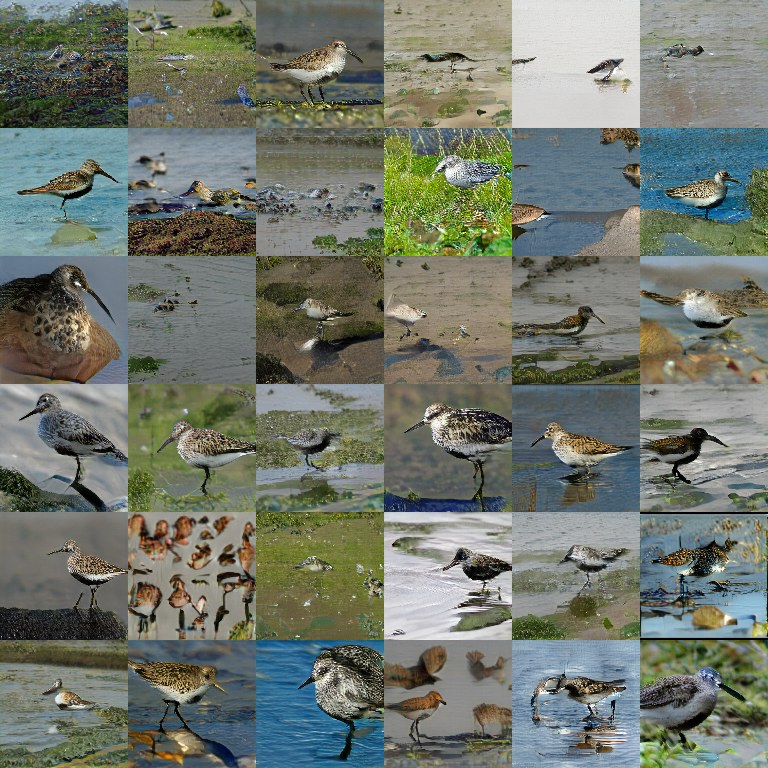}
  \caption{SAGAN IFID 287.50}
\end{subfigure}%
\begin{subfigure}{.2\textwidth}
  \centering
  \includegraphics[width=.98\linewidth]{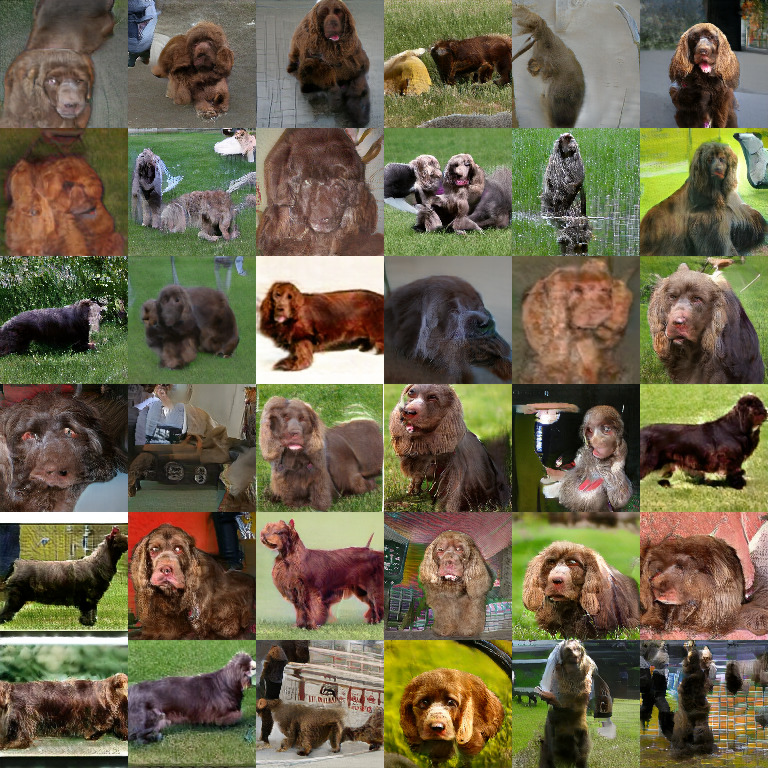}
  \caption{SAGAN IFID 275.03}
\end{subfigure}%
\begin{subfigure}{.2\textwidth}
  \centering
  \includegraphics[width=.98\linewidth]{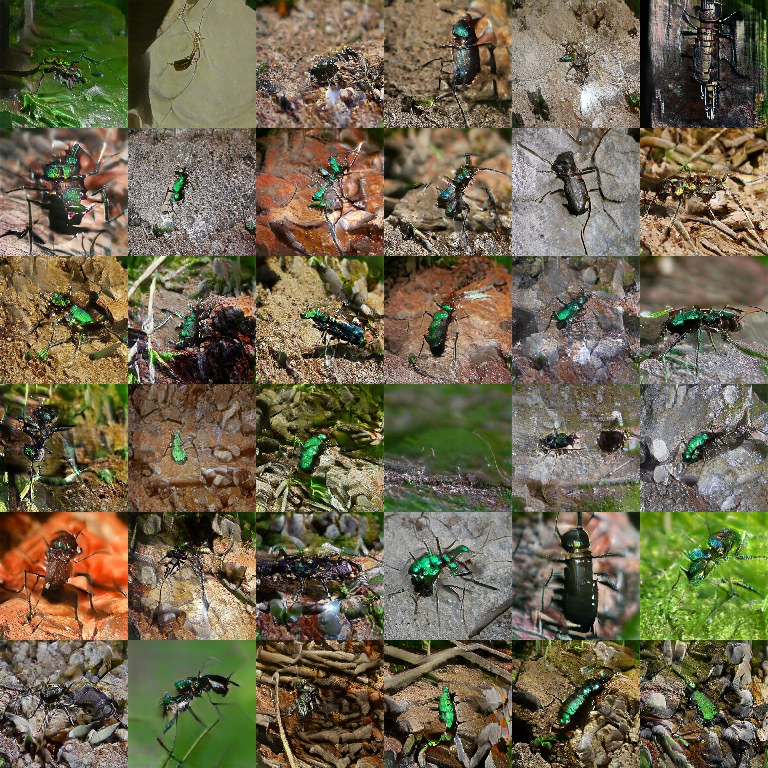}
  \caption{SAGAN IFID 339.48}
\end{subfigure}%
\begin{subfigure}{.2\textwidth}
  \centering
  \includegraphics[width=.98\linewidth]{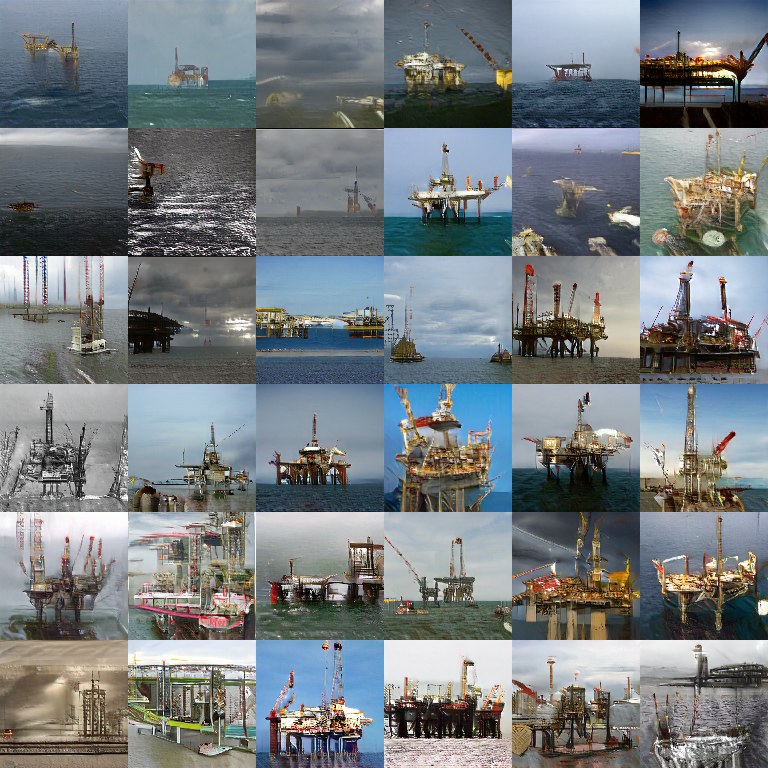}
  \caption{SAGAN IFID 307.16}
\end{subfigure}%
\begin{subfigure}{.2\textwidth}
  \centering
  \includegraphics[width=.98\linewidth]{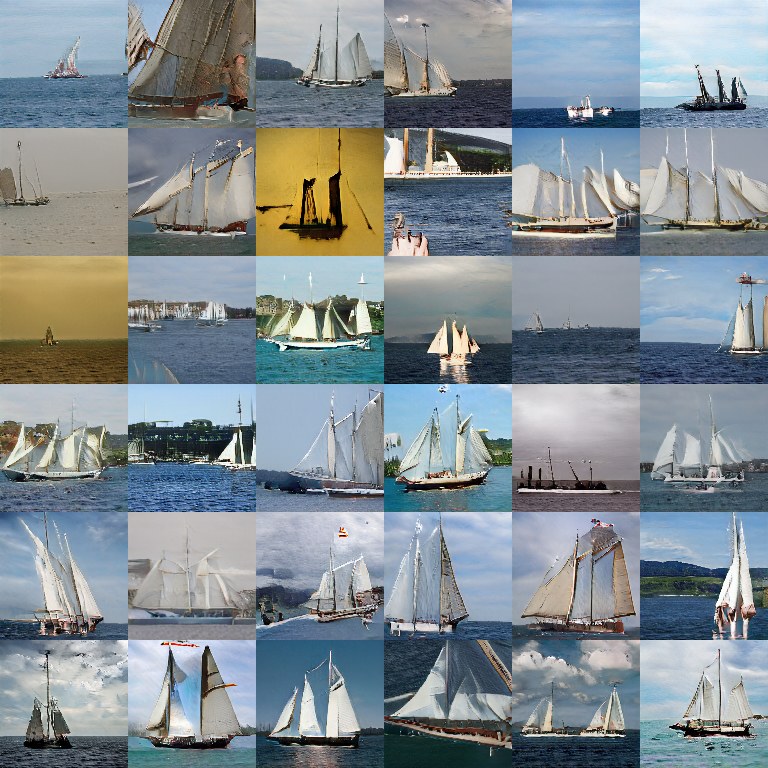}
  \caption{SAGAN IFID 296.28}
\end{subfigure}
\hfill
   \caption{ Classes for which Intra-FID gets better for our proposed MHGAN. We show random samples from the 5 classes that improved the most in \Cref{fig:I128_IFID} by points, excluding the trout class where SAGAN completely collapses.}
\label{fig:IFID_better}
\end{figure*}
\begin{figure*}[t]
\centering
\begin{subfigure}{.2\textwidth}
  \centering
  \includegraphics[width=.98\linewidth]{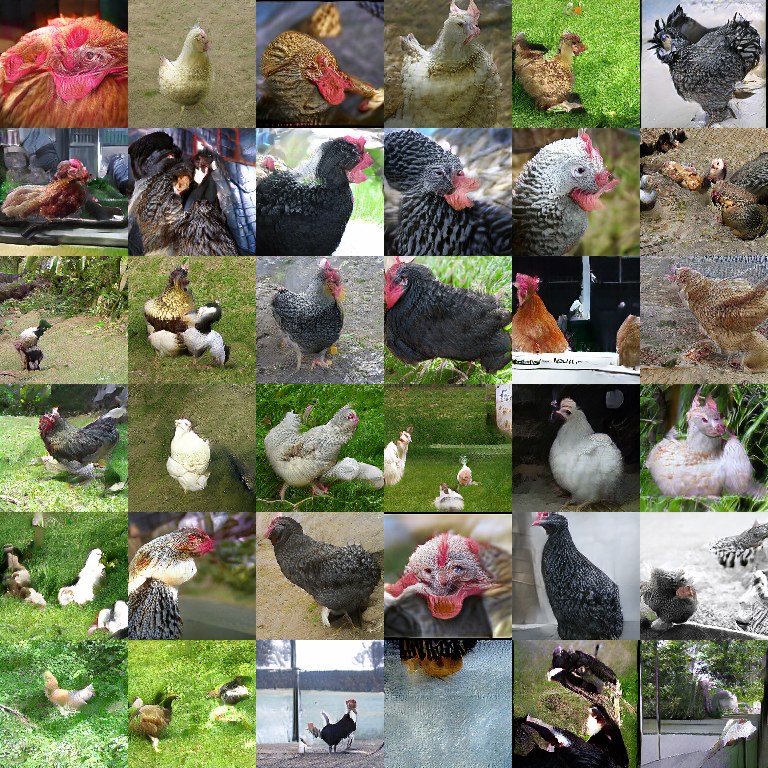}
  \caption{MHGAN IFID 27.92}
\end{subfigure}%
\begin{subfigure}{.2\textwidth}
  \centering
  \includegraphics[width=.98\linewidth]{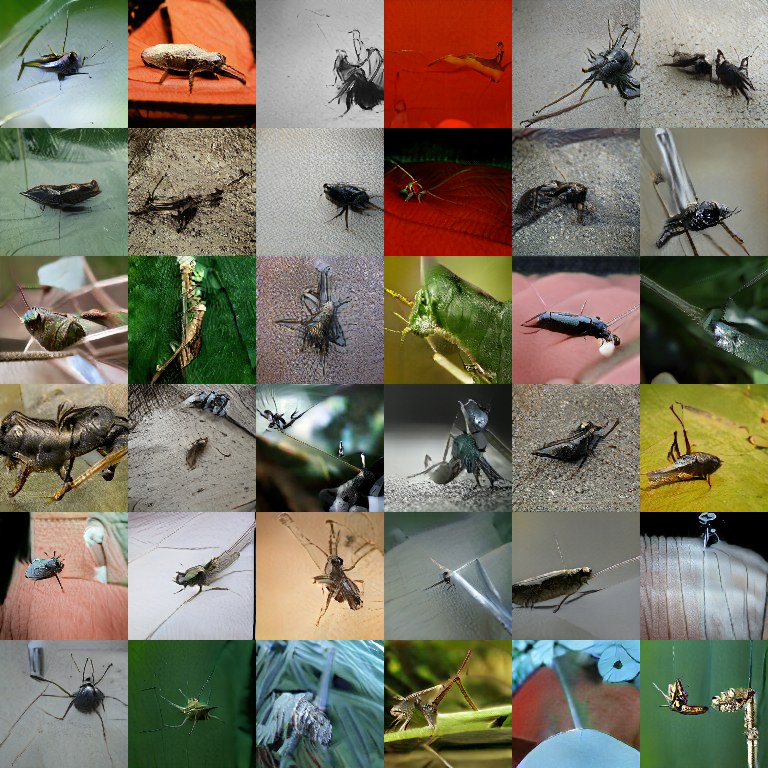}
  \caption{MHGAN IFID 93.79}
\end{subfigure}%
\begin{subfigure}{.2\textwidth}
  \centering
  \includegraphics[width=.98\linewidth]{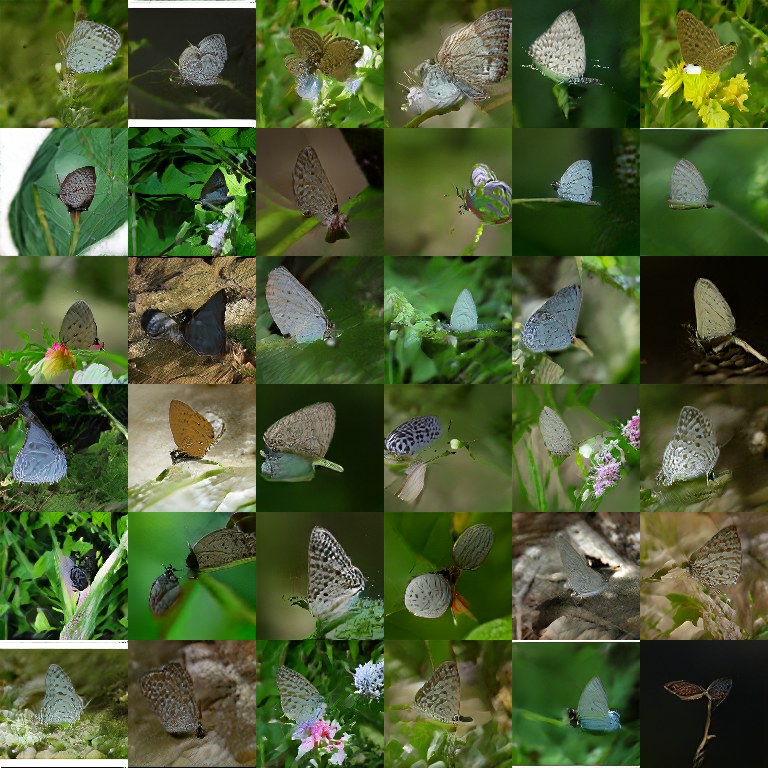}
  \caption{MHGAN IFID 154.82}
\end{subfigure}%
\begin{subfigure}{.2\textwidth}
  \centering
  \includegraphics[width=.98\linewidth]{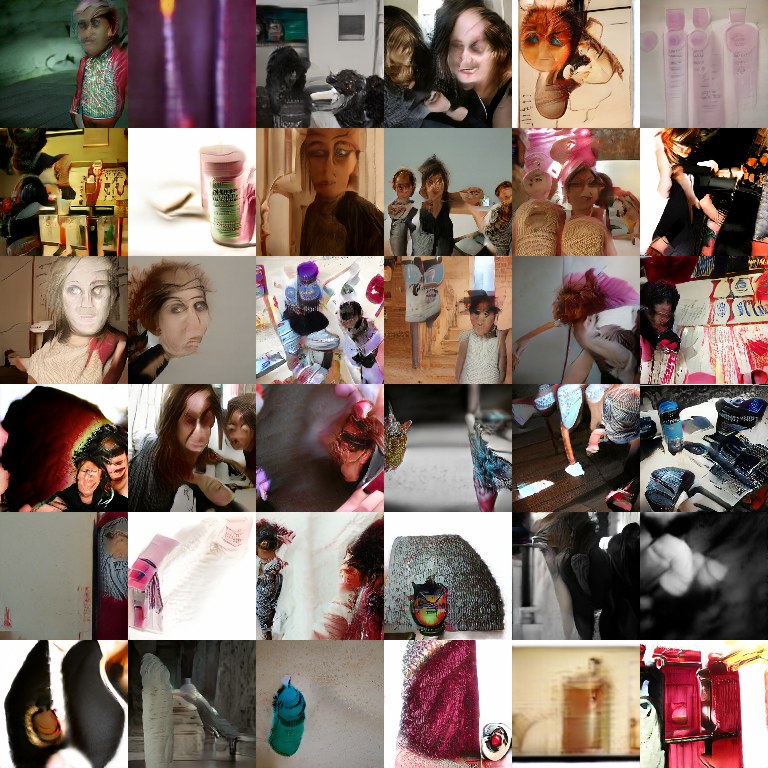}
  \caption{MHGAN IFID 102.92}
\end{subfigure}%
\begin{subfigure}{.2\textwidth}
  \centering
  \includegraphics[width=.98\linewidth]{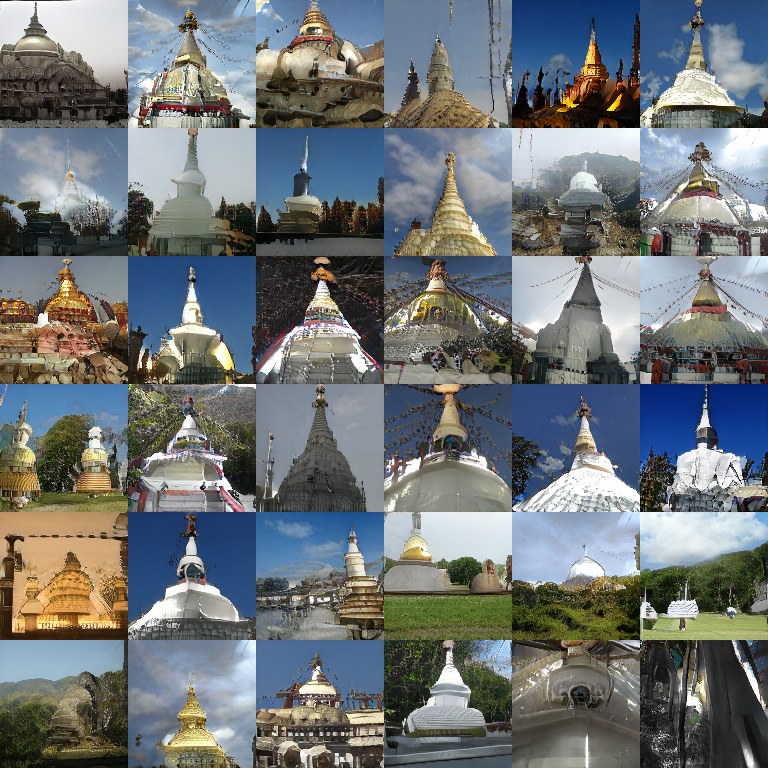}
  \caption{MHGAN IFID 182.66}
\end{subfigure}
\hfill
\begin{subfigure}{.2\textwidth}
  \centering
  \includegraphics[width=.98\linewidth]{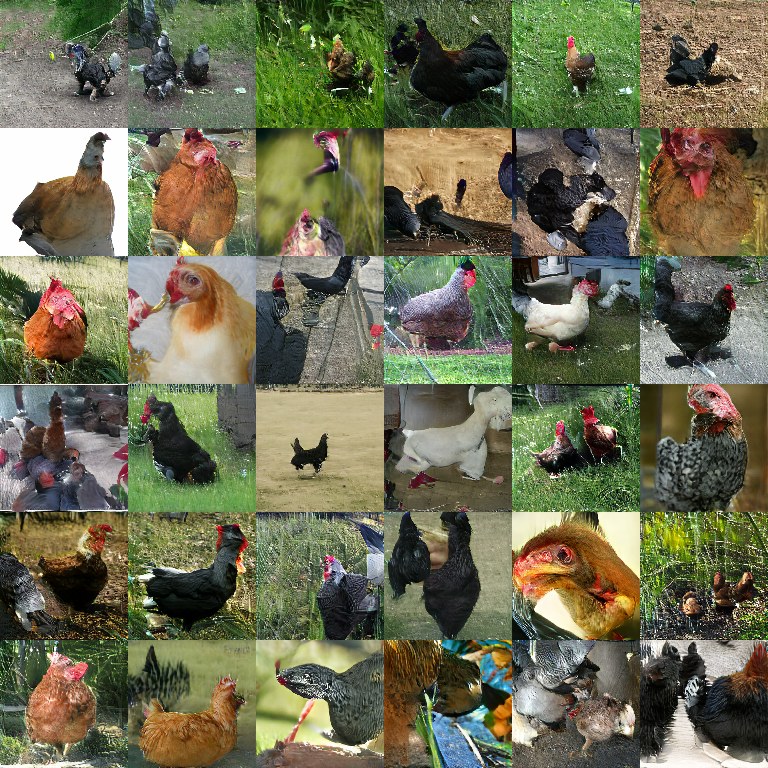}
  \caption{SAGAN IFID 19.48}
\end{subfigure}%
\begin{subfigure}{.2\textwidth}
  \centering
  \includegraphics[width=.98\linewidth]{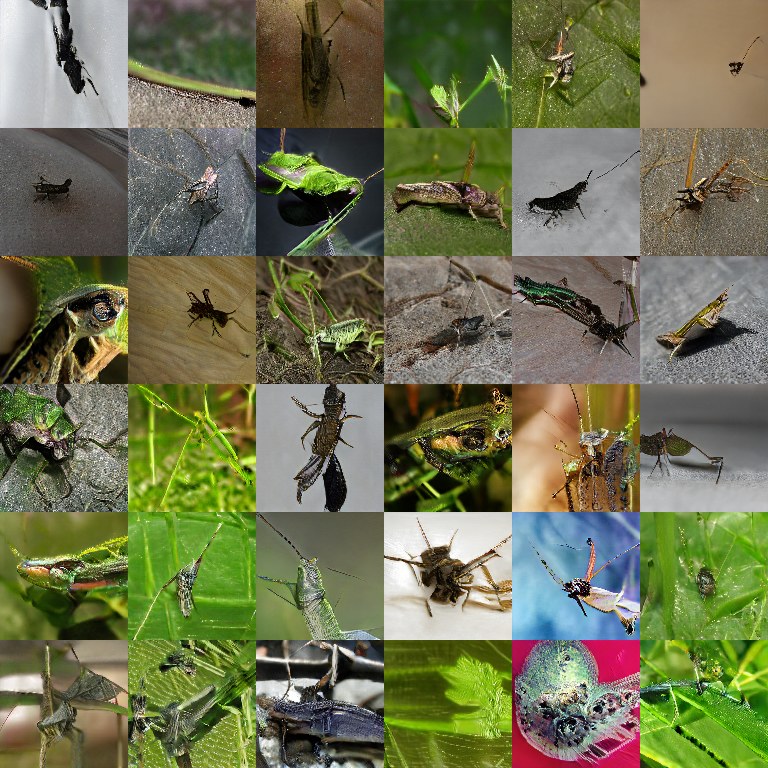}
  \caption{SAGAN IFID 81.93}
\end{subfigure}%
\begin{subfigure}{.2\textwidth}
  \centering
  \includegraphics[width=.98\linewidth]{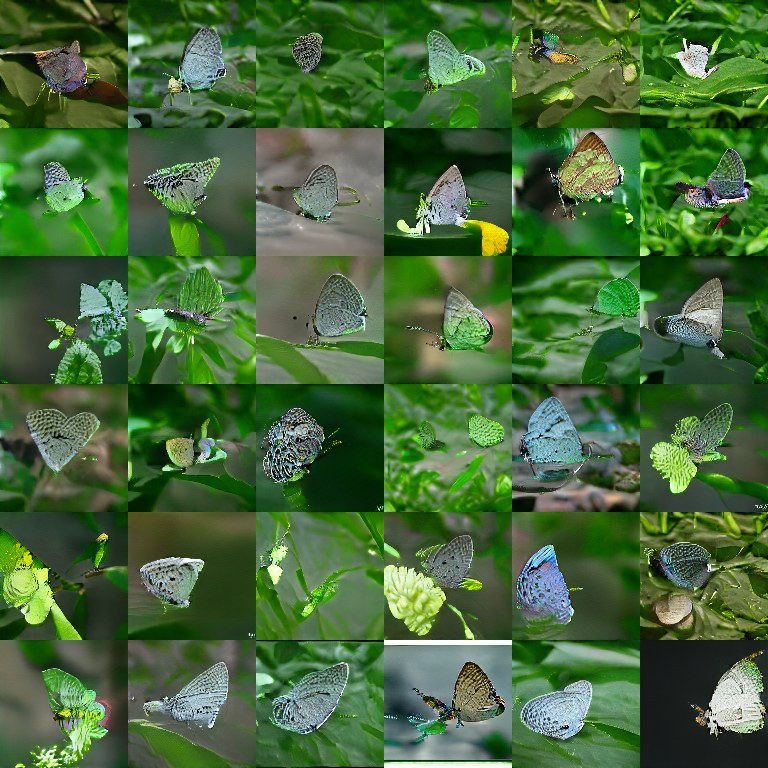}
  \caption{SAGAN IFID 146.08}
\end{subfigure}%
\begin{subfigure}{.2\textwidth}
  \centering
  \includegraphics[width=.98\linewidth]{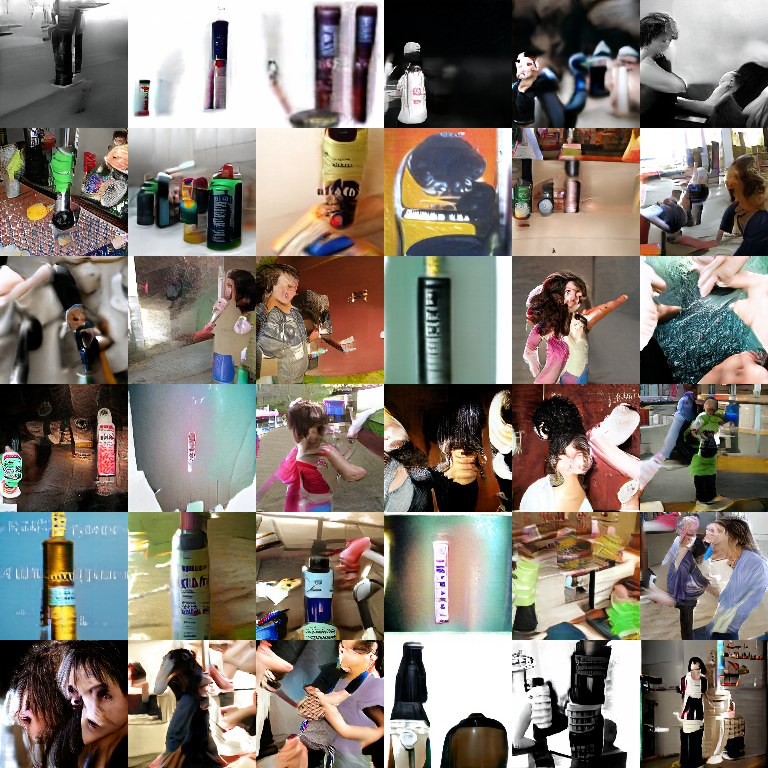}
  \caption{SAGAN IFID 94.21}
\end{subfigure}%
\begin{subfigure}{.2\textwidth}
  \centering
  \includegraphics[width=.98\linewidth]{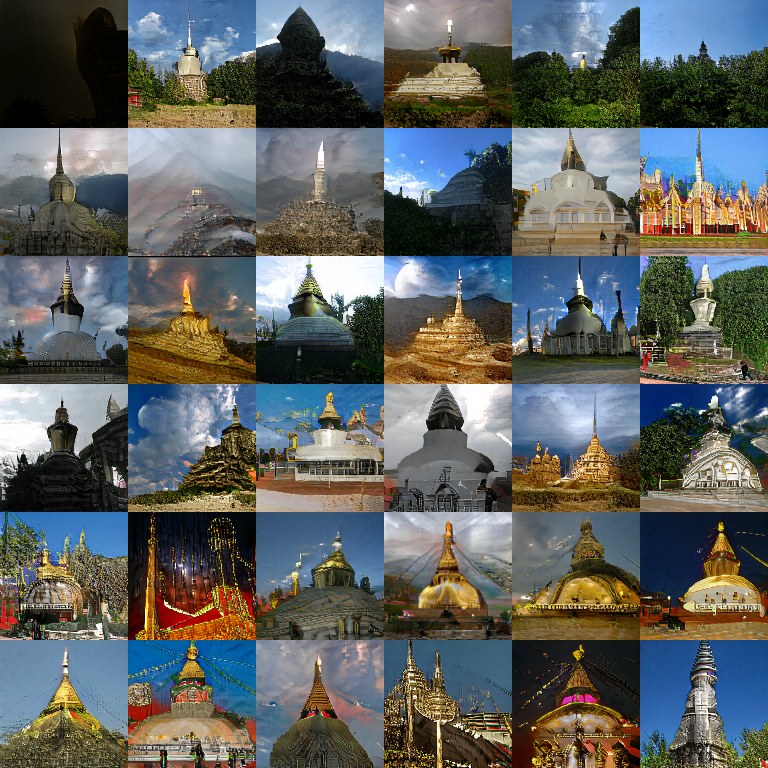}
  \caption{SAGAN IFID 173.95}
\end{subfigure}
\hfill
   \caption{ Classes for which Intra-FID gets worse for our proposed MHGAN. We show random samples from the 5 classes that worsened the most in \Cref{fig:I128_IFID} by points. We did not observe any instances of mode collapse in MHGAN.}
\label{fig:IFID_worse}
\end{figure*}
For SAGAN, MHGAN, and MHGAN-SSL we optimize with size 1024 batches, learning rates of $1e-4$ and $4e-4$ for $G$ and $D$, and 1 $D$ step per $G$ step.
For ACGAN and ACGAN-SSL, we found training to be unsuccessful with only 1 $D$ step per $G$ step, so we train with 2 $D$ steps per $G$ step (training with more than 2 $D$ steps did not improve results).
For ACGAN-SSL we also used a learning rate of $5e-4$ for $D$ and a $z$ dimension of 120 instead of 128 in our other experiments \cite{lucic_ssl}.
The generator's auxiliary classifier loss weight is fixed at $\lambda=0.1$ for all experiments.
We use 64 channels and limit most of our experiments to 1,000,000 iterations.
We use the Inception Score (IS) \cite{salismans} and Frechet Inception Distance (FID) for quantitative evaluation of our generative models. During training we compute these scores with 10 groups of 1024 randomly generated samples using the official tensorflow implementations \cite{tfgan,ISFID_python}, and for the final numbers in \Cref{tab:i128scores} we use 50k samples.
%
%
\subsection{Fully supervised image generation}
In \Cref{fig:perf_charts} and \Cref{tab:i128scores} we present fully supervised results on the Imagenet dataset \cite{imagenet_data}.
Previous work has noted a multitude of GAN algorithms that train well on datasets of limited complexity and resolution but may not provide an indication that they can scale \cite{lucic}.
Thus we choose the largest and most diverse image data set commonly used to evaluate our GANs.
Imagenet contains 1.3M training images and 50k validation images, each corresponding to one of 1k object classes. We resize the images to $128\times 128$ for our experiments.
On a single v3-8 TPU MHGAN and SAGAN complete 10k steps every 2 hours, and ACGAN completes 10k G-steps every 3.3 hours. Thus 1M MHGAN iterations takes 8.3 days.

\Cref{tab:i128scores} shows that the auxiliary MHGAN loss added to SAGAN trains a better GAN according to Inception score and FID.
The classifier increases the fidelity of the samples generated by the GAN without sacrificing diversity as shown in \Cref{fig:I128_IFID}, where we plot the intra-class FID of SAGAN versus MHGAN for the best models we trained by overall IS.
The mean class FID improves by 3.5\%, and the class FID is lowered for 622 classes, and for 46 classes in the lower right cluster improves by an average of 50\%.
For the point below the cluster, mode collapse is prevented by MHGAN for the tench class.
Some of these points in the cluster are shown in \Cref{fig:IFID_better}, where we randomly sample images from classes where the intra-FID score decreased (improved) from the SAGAN baseline to our MHGAN.
We see in some of these cases that SAGAN is showing distortions or early signs of mode collapse, despite overall IS being at a high.
We also show the opposite relation in \Cref{fig:IFID_worse}, where we randomly sample the classes for which FID increases (worsens) the most from SAGAN to MHGAN.
In \Cref{fig:IFID_worse} we see that MHGAN doesn't show a worrying level of decreased diversity or mode collapse.
\begin{table}[!htb]
      \centering
        \begin{tabular}{lcc}
            \toprule
            \multicolumn{1}{c}{Method} &
            \multicolumn{2}{c}{\textbf{Imagenet-128}} \\
            
            \multicolumn{1}{c}{}    & 
            IS & FID       \\
            \midrule
            Real data  & 156.40  \\
            \midrule
            SAGAN                & 52.79 & 16.39    \\
            ACGAN                & 48.94 & 24.72    \\
            MHGAN                 & {\bf 61.98} & {\bf 13.27}    \\
            \midrule
            SAGAN 1M \cite{SAGAN}  & 52.52 & 18.65 \\
            BigGAN 1M \cite{BigGAN}  & 63.03 & 14.88  \\
            \bottomrule
        \end{tabular}
    \caption{Inception Scores and FIDs for supervised image generation on Imagenet-128. The models were chosen by maximizing the IS within 1M iterations. The BigGAN comparison we include is the one most similar to our setting (batch size 1024). Our SAGAN Baseline results are consistent with the results reported online with the implementation we use \cite{tfgan}. }
    \label{tab:i128scores}
\end{table}
\subsection{Measuring the conditioning of the discriminator and generator}
\label{sec:classspec}
We attribute our model's success to the fact that it gradually incorporates class specific information into both the discriminator and the generator networks than just projection discrimination alone.
When we have a classifier as in MHGAN and ACGAN it is straightforward to calculate the {\it validation accuracy} and {\it self accuracy}.
Validation accuracy is the accuracy of the classifier on real validation data, the whole standard validation partition of the dataset is used. This measures how good the classifier learned is; if it starts to overfit the training data then we expect validation accuracy to decay.
For self accuracy we test if $\arg\max_{k\in\mathcal{Y}} C_k(G(z,y))$ equals $y$. This measures how G incorporates the label information into its output, as measured by the concurrently trained C. This measures the self consistency of the GAN.
For the baseline SAGAN network with a projection discriminator, it is also possible to perform classification using the method mentioned in \Cref{sec:mot}.
That is our "classification" for an example $x$ is $\arg\max_{k\in\mathcal{Y}} D(x,k)$, where each $k$ is used as input to the projection discriminator layer that was trained.

\medskip
\noindent
{\bf Conditional information in C\&D.}
We find that our suspicion from \Cref{sec:mot} is confirmed for the baseline projection discriminator model: for $(x,y)\sim\pdata$ it is not the case that $D(x,y)>D(x,k), k\neq y$ with high probability.
In fact, we find that for all projection discriminators in our trained models that $D(x,y)>D(x,k), k\neq y$ uniformly and randomly.

\begin{figure}[t]
\begin{subfigure}{.27\textwidth}
  \centering
  \includegraphics[width=.98\linewidth]{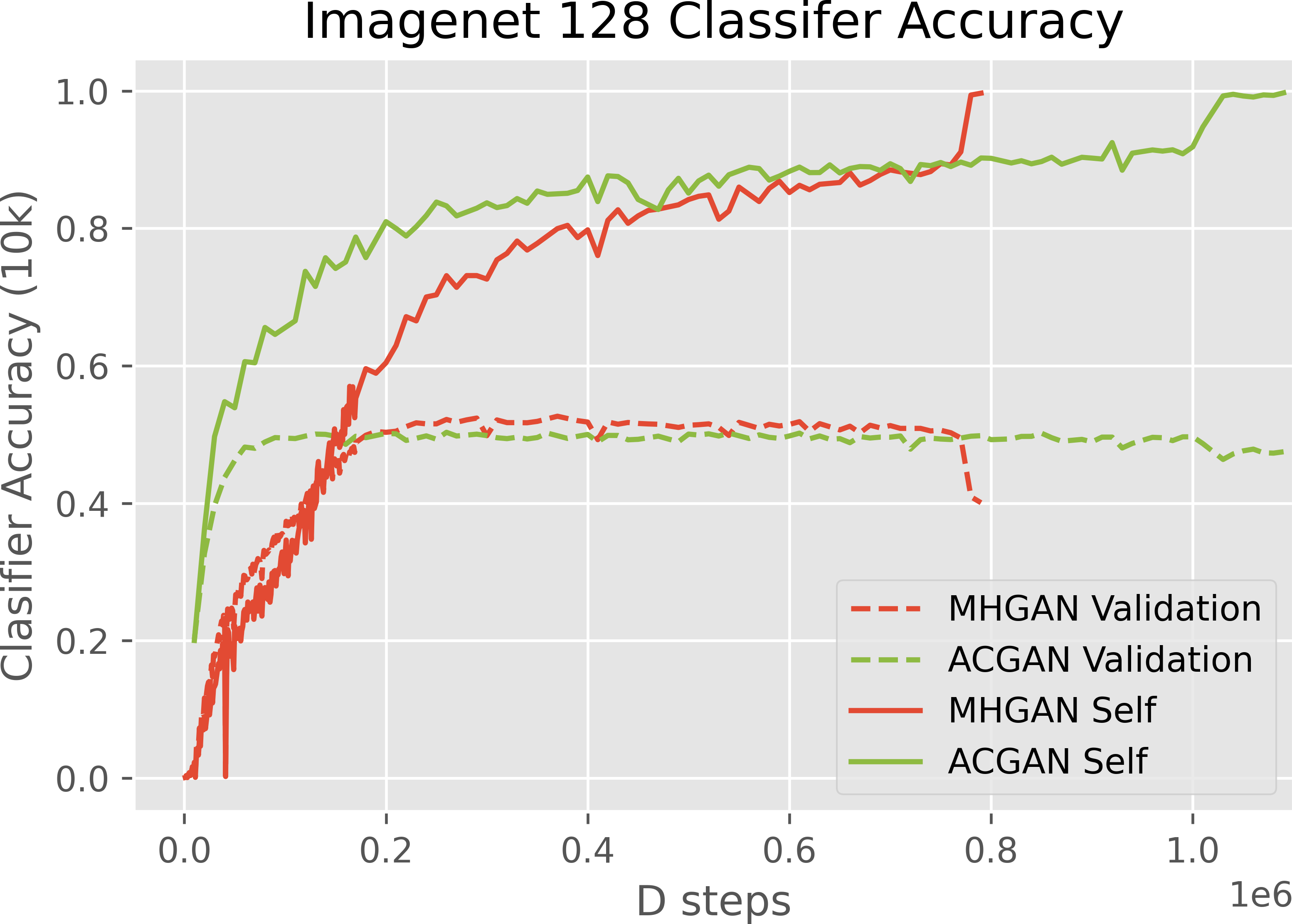}
  \caption{}
  \label{fig:i128_accs}
\end{subfigure}%
\begin{subfigure}{.19\textwidth}
  \centering
  \includegraphics[width=.98\linewidth]{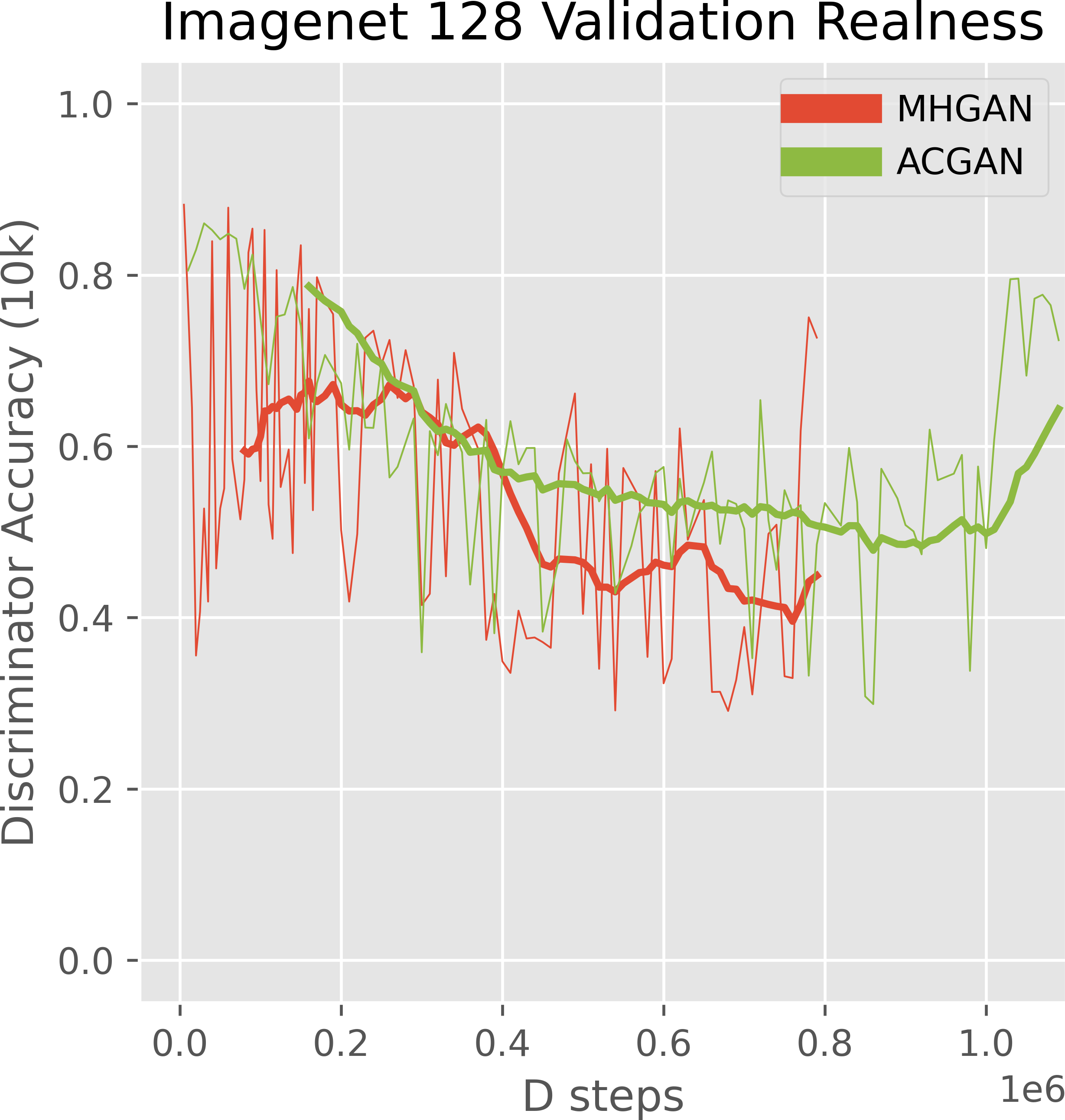}
  \caption{}
  \label{fig:i128_dval}
\end{subfigure}
   \caption{Validation accuracy and self accuracy track IS and FID improvements and reach their own plateaus during training. Discriminator accuracy is very noisy during training, and behaves differently than classification accuracy, suggesting that the two tasks can be separately optimized.}
\label{fig:aux_metrics}
\end{figure}

However, the motivation behind \cref{eq:mhsimplest} was to create a better GAN by training the discriminator to incorporate as much conditional information in the dataset.
Though we find that the projection discriminator layer cannot be made useful for classification, we can still incorporate more conditional information into the discriminator network through the embeddings it produces.
In \Cref{fig:i128_accs} we plot validation and self accuracy for the models with a classifier (these metrics oscillate randomly near "$1/$ num classes" throughout training for SAGAN).

We see that validation performance (top-1 accuracy on Imagenet) is about the same for both models, converging around 50\% and peaking at 50.32\% for ACGAN and 52.66\% for MHGAN. This is not competitive with purpose built classifiers which have different architectures and are typically much deeper than the discriminator of our GANs.
Meanwhile classification accuracy on the training set hovers around the level of 98\%.
Self accuracy converges to 90\% in both, and for obvious reasons goes to 100\% when the generator experiences collapse (it is a curious detail that the SOA top-1 classification accuracy on Imagenet has also been converging to 90\% \cite{imagenet_soa}).
We leave it to future work to control self accuracy to be at the same level as validation accuracy, and for both to increase even more gradually, since this could prolong training and IS and FID improvements further.
Intuitively there is a trade-off between fidelity (generated images looking like the right class) and diversity.
In \Cref{fig:i128_accs} we see that the absence of a logarithm, and the hinge function which limits learning on examples that are correct by a margin, leads to the more gradual incorporation of class specific information into the generator for MHGAN.

In \Cref{fig:i128_dval} we plot the discriminator's accuracy on the real validation set (the percentage of examples for which $D(x,y)>0$). This is a very noisy metric during training, so we also plot the 100k itr moving average with the heavier line.
It is interesting that this metric declines to around 50\% as training progresses, since the same discriminator accuracy on the training set is stable at 91\% on the training set for MHGAN, and 98\% for SAGAN.
Intuitively this indicates that $D$ is memorizing the training set \cite{BigGAN}.
Yet since the classification accuracy remains stable or declines only slightly, this shows a disconnect between the classification and discrimination tasks.

\begin{figure}[t]
\begin{subfigure}{.23\textwidth}
  \centering
  \includegraphics[width=1\linewidth]{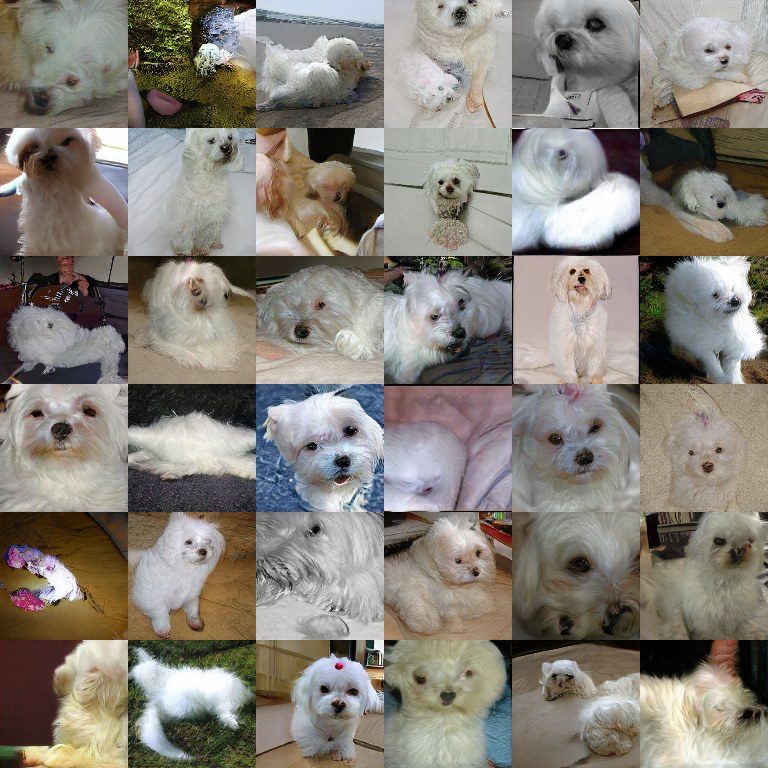}
\end{subfigure}%
\hfill
\begin{subfigure}{.23\textwidth}
  \centering
  \includegraphics[width=1\linewidth]{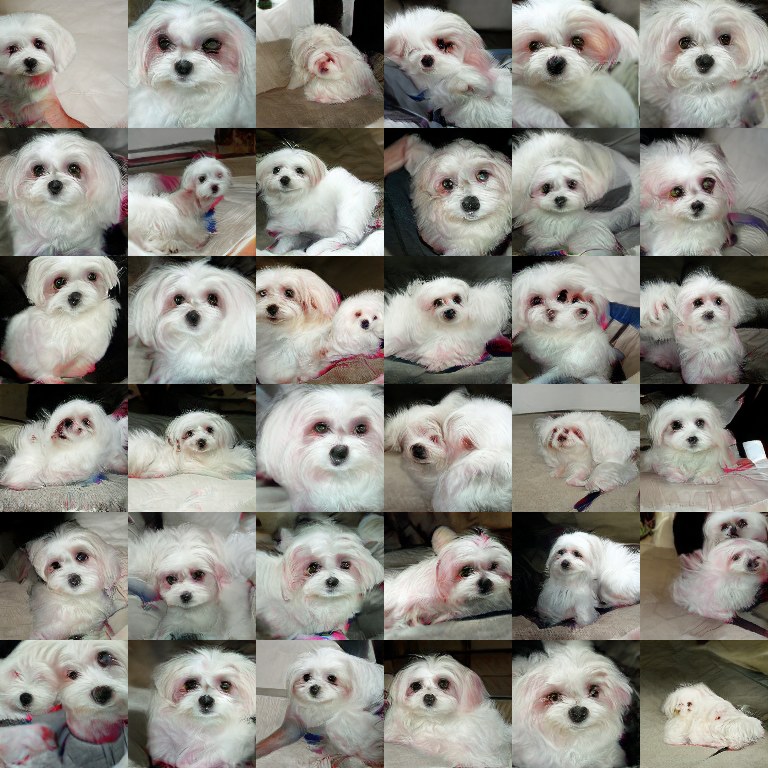}
\end{subfigure}
\begin{subfigure}{.23\textwidth}
  \centering
  \includegraphics[width=1\linewidth]{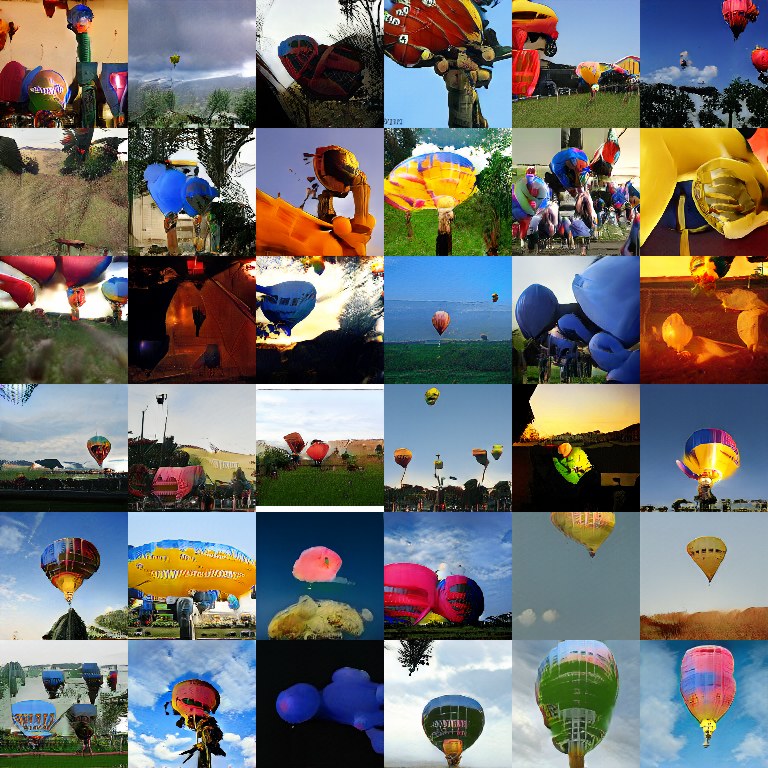}
  \caption{SAGAN trained for 580k steps. IS 47.79 and FID 17.10 (50k) at this stage.}
\end{subfigure}%
\hfill
\begin{subfigure}{.23\textwidth}
  \centering
  \includegraphics[width=1\linewidth]{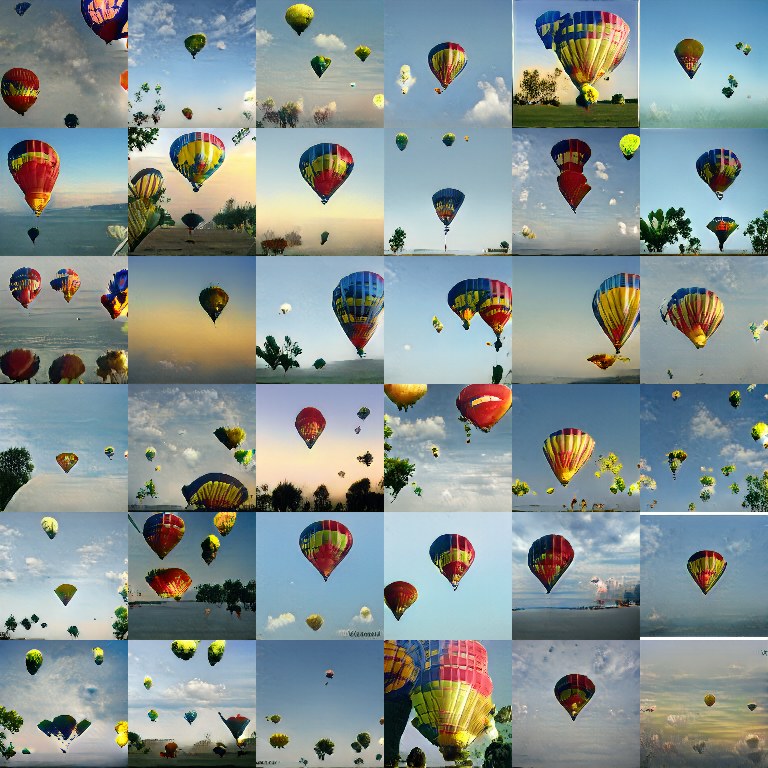}
  \caption{The model after 5k steps of MHSharedGAN training. IS 169.68 and FID 8.87 (50k).}
\end{subfigure}
\caption{Enforcing class fidelity without an auxiliary classifier, and training class margins to be respected in the projection discriminator using \cref{eq:mhsimplest} instead leads to low diversity, but higher quality images.}
\label{fig:low_div}
\end{figure}

\medskip
\noindent
{\bf The importance of separating classification and discrimination tasks.}
Earlier we mentioned that for all three models that we compare, the projection discriminator never incorporates class specific information; that the discriminator output has the highest affinity for the correct class about "$1/$ num classes" of the time.
We observe that it is disadvantageous to try and train the projection discriminator to have the property $D(x,y)>D(x,k), k\neq y$ with high probability.
Both MHGAN and ACGAN use an extra fully connected layer as an auxiliary classifier during training. Having this extra layer and training with \cref{eq:overallhinge} improves both quality and diversity as shown in \Cref{fig:perf_charts}.
It is also possible to follow the approach mentioned in \Cref{sec:mot} and share parameters between the projection discriminator and the classifier, and train with \cref{eq:overallhinge}. We call this MHSharedGAN and illustrate this strategy in \Cref{fig:MHSharedGANarch}.

Training from scratch this way does not lead to satisfactory performance, but introducing this loss in the middle of SAGAN training produces interesting results (introducing it after 1M steps is also unsuccessful).
After just 5k iterations after transitioning from training with \cref{eq:orighinge} (SAGAN) to \cref{eq:overallhinge} (MHSharedGAN) Inception score skyrockets from 47.79 to 169.68 and Frechet Inception Distance drops to 8.87 (evaluated on 50k images).
The catch is that unlike with MHGAN, there is a trade-off between quality and diversity being made when parameters are shared.
This is illustrated in \Cref{fig:low_div}, where we see the diversity of the generator drops drastically.
Though the layer is spectrally normalized during training with 1 step of the power iteration method, during this second phase of training the spectrum of the projection discriminator (rank v.s. value plot of eigenvalues) changes from a sigmoid shape to a steep exponential shape, indicating that the projection has collapsed to a just a few dimensions.

In developing MHGAN we also experimented with $K+1$ \cite{salismans} formulations of the loss where the discrimination and classification tasks are unified. At low resolutions ($48\times48$ and below) we observed that such models are able to train from scratch and obtain competitive IS and FID scores, but that fidelity came at the cost of diversity there too. 
Our experiments show that classification should be left as an auxiliary task to discrimination, and not combined with it by sharing parameters too closely.

\subsection{Semi-supervised image generation}
To demonstrate our multi-hinge loss in a semi-supervised setting we use the partially labeled Imagenet data set with a random selection of 10\% of the samples from each class retaining their label publicly available on TFDS \cite{tfds}.
We perform experiments on this $128\times128$ sized dataset.
We keep the same architecture choices described in \Cref{sec:exp} and train our MHGAN-SSL with \cref{eq:overallhingessl} and compare to ACGAN-SSL trained with \cref{eq:acssl}.
MHGAN-SSL trains faster than ACGAN-SSL but reaches a similar level of performance, with (IS, FID) scores of (32.40, 26.12) for MHGAN-SSL and (32.38, 26.12) for the ACGAN-SSL baseline. Both networks have their validation accuracy go to 40\%, and self accuracy above 80\%.
The speed of of MHGAN-SSL is an advantage over the ACGAN-SSL we train, which has a similar loss formulation to S\textsuperscript{2}GAN-CO which achieves (IS, FID) scores of (37.2, 17.7) using a larger architecture than we train \cite{lucic_ssl}.
We leave it to future work to train the network in S\textsuperscript{2}GAN-CO with the MHGAN-SSL loss.
However it has been shown that co-training with pseudo-labels is not as competitive as pre-labeling the unlabeled data with a separate classifier and then using a fully supervised ACGAN like loss, this approach S\textsuperscript{2}GAN achieves (IS, FID) scores of (73.4, 8.9)  \cite{lucic_ssl}.
We also leave it to future work to use this pre-labeling strategy with MHGAN to improve on SSL GAN training.

\section{Conclusion}
MHGAN is a powerful addition to projection discrimination and improves training with negligible additional computational cost. \Cref{tab:i128scores} and \Cref{fig:perf_charts} show that the multi-hinge loss improves both the quality and diversity of generated images on Imagenet-128.
We also show how classification and discrimination tasks should not be integrated too closely, and show how the multi-hinge loss can be used to trade diversity for image quality in MHSharedGAN.
MHGAN is able to perform well in both fully supervised and semi-supervised settings, and learns a relatively accurate classifier concurrently with a high quality generator.

\section*{Acknowledgements}
Ilya Kavalerov and Rama Chellappa acknowledge the support of the MURI from the Army Research Office under the Grant No. W911NF-17-1-0304. This is part of the collaboration between US DOD, UK MOD and UK Engineering and Physical Research Council (EPSRC) under the Multidisciplinary University Research Initiative.
Wojciech Czaja is supported in part by LTS through Maryland Procurement Office and the NSF DMS 1738003 grant.
The authors thank Tensorflow Research Cloud (TFRC) and Google Cloud Platform (GCP) for their support.

\newpage

{\small
\bibliographystyle{ieee_fullname}
\bibliography{egbib}
}

\clearpage

\section{Appendix: Source Code}

Our MHGAN and baseline implementations are based on the Self-Attention GAN \cite{SAGAN} available on github \cite{tfgan}. Our code is available at \url{https://github.com/ilyakava/gan}.

\section{Appendix: Additional details on our SSL GANs}

\begin{figure*}[htb!]
	\centering
	\begin{subfigure}{.5\textwidth}
		\centering
		\includegraphics[width=1\linewidth]{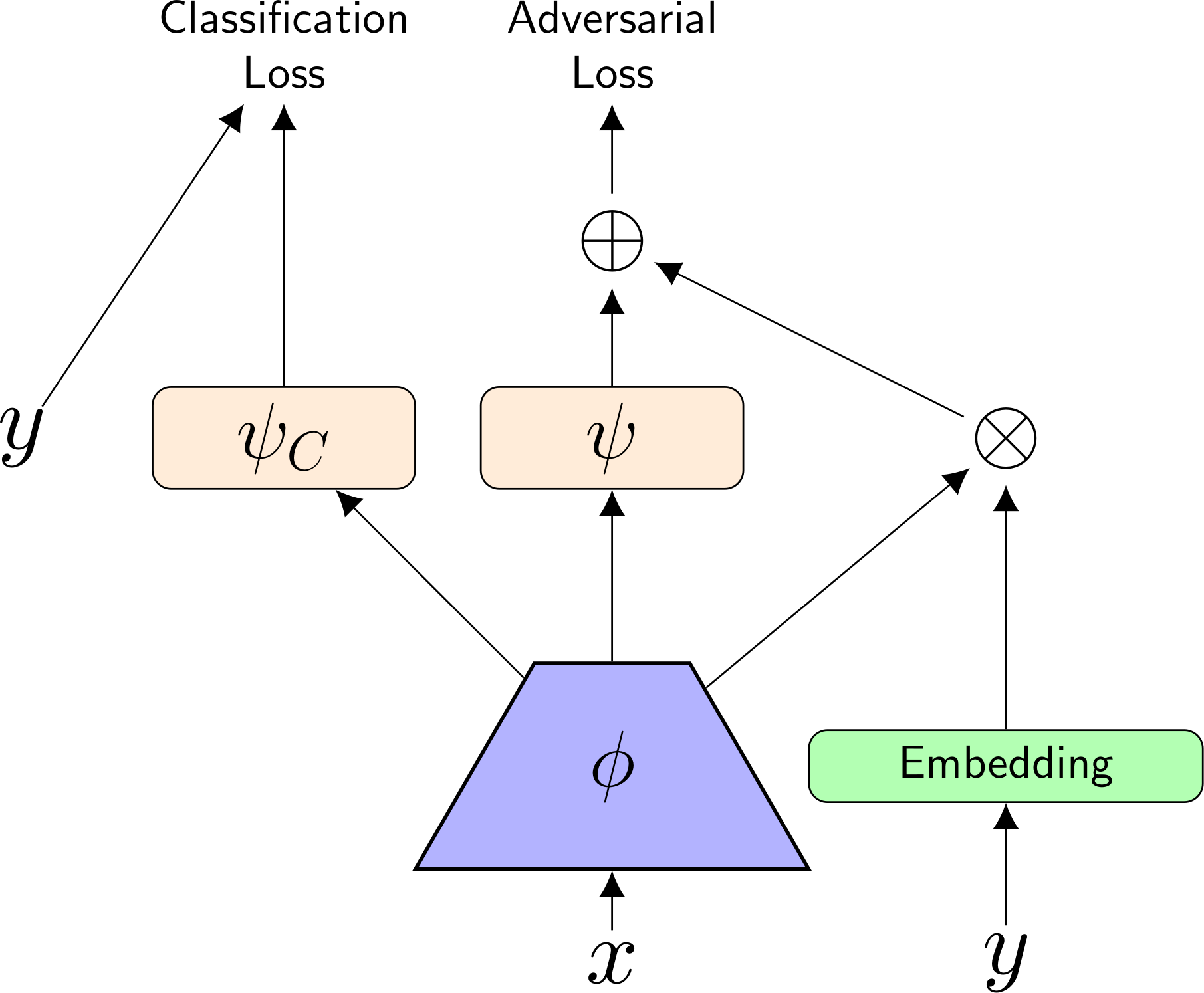}
		\caption{For $x\sim\pdata$}
		\label{fig:xlab}
	\end{subfigure}%
	\hfill
	\begin{subfigure}{.5\textwidth}
		\centering
		\includegraphics[width=.7\linewidth]{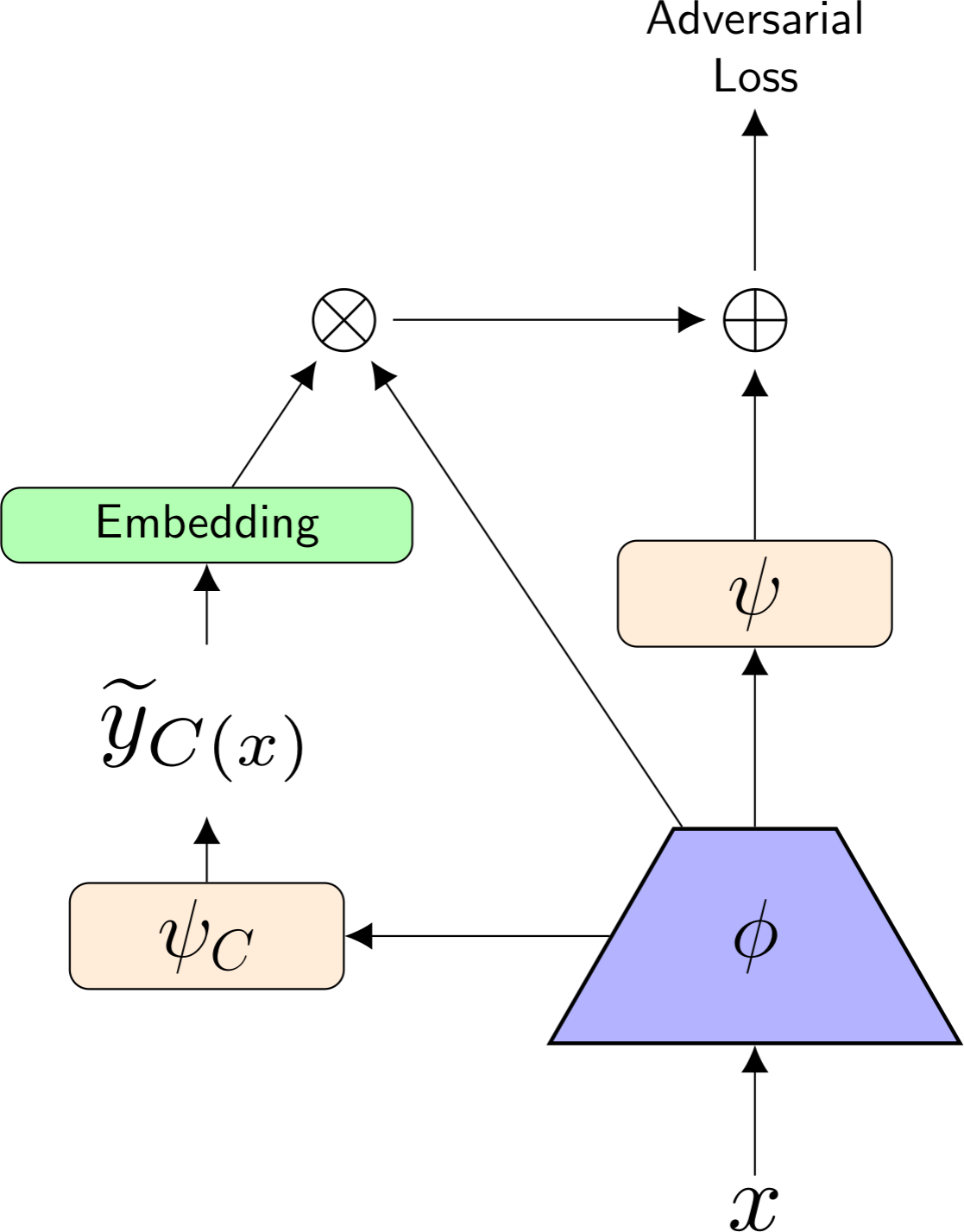}
		\caption{For $x\sim\punlab$}
		\label{fig:xunlab}
	\end{subfigure}
	\caption{The architectures of the semi supervised GANs that we train. \ref{fig:xlab} shows the auxiliary classifier architecture \cite{projdisc} of our SAGAN, the same as used for fully supervised experiments. When $x\sim\pdata$ the discriminator treats the example as shown in \ref{fig:xlab}. In the diagram the green "Embedding" block is the projection discriminator embedding for a class, $\phi$ is the discriminator up to the penultimate feature layer, $\psi$ is a linear layer with scalar output, $\psi_C$ is a linear layer with output size $n$ classes.  When $x\sim\punlab$ and the label is not available, it is treated as shown in \ref{fig:xunlab}. This is similar to S\textsuperscript{2}GAN \cite{lucic_ssl}. Both ACGAN-SLL and MHGAN-SLL use these architectures, the only difference between the two is the form of the classification loss for labelled examples.}
	\label{fig:arches}
\end{figure*}

\Cref{fig:arches} shows how labelled and unlabelled data flow through our ACAN and MHGAN networks during semi-supervised learning.
In \Cref{fig:xlab} is a projection discrimination network with an auxiliary classifier added.
\Cref{fig:xunlab} shows that for unlabeled data, the role of $y$ is substituted with the pseudolabel $\widetilde{y}_{C(x)}$, and the classification loss is not trained.

\begin{figure*}[htb!]
	\centering
	\begin{subfigure}{.5\textwidth}
		\centering
		\includegraphics[width=1\linewidth]{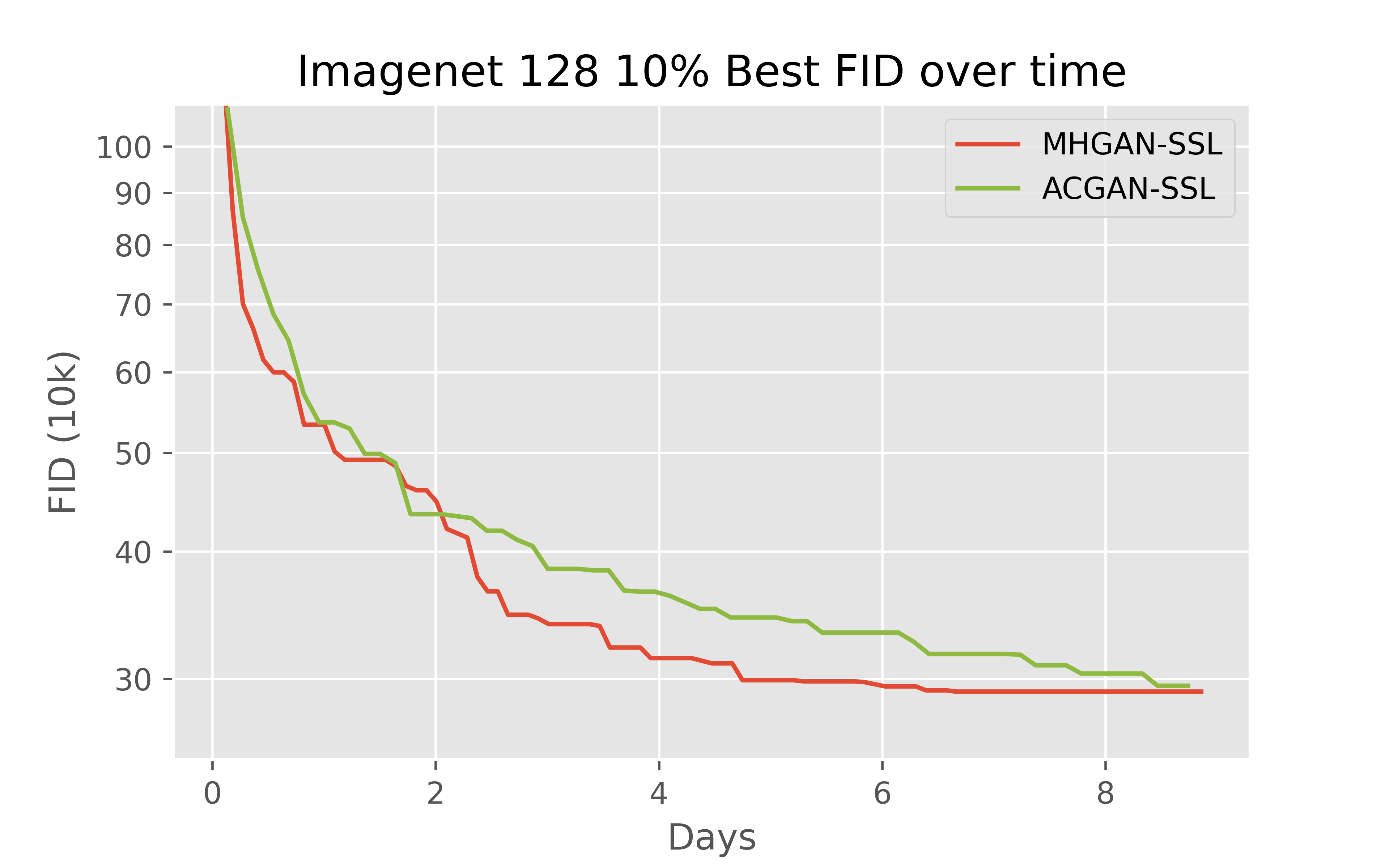}
		\caption{Semi-Supervised}
	\end{subfigure}%
	\hfill
	\begin{subfigure}{.5\textwidth}
		\centering
		\includegraphics[width=1\linewidth]{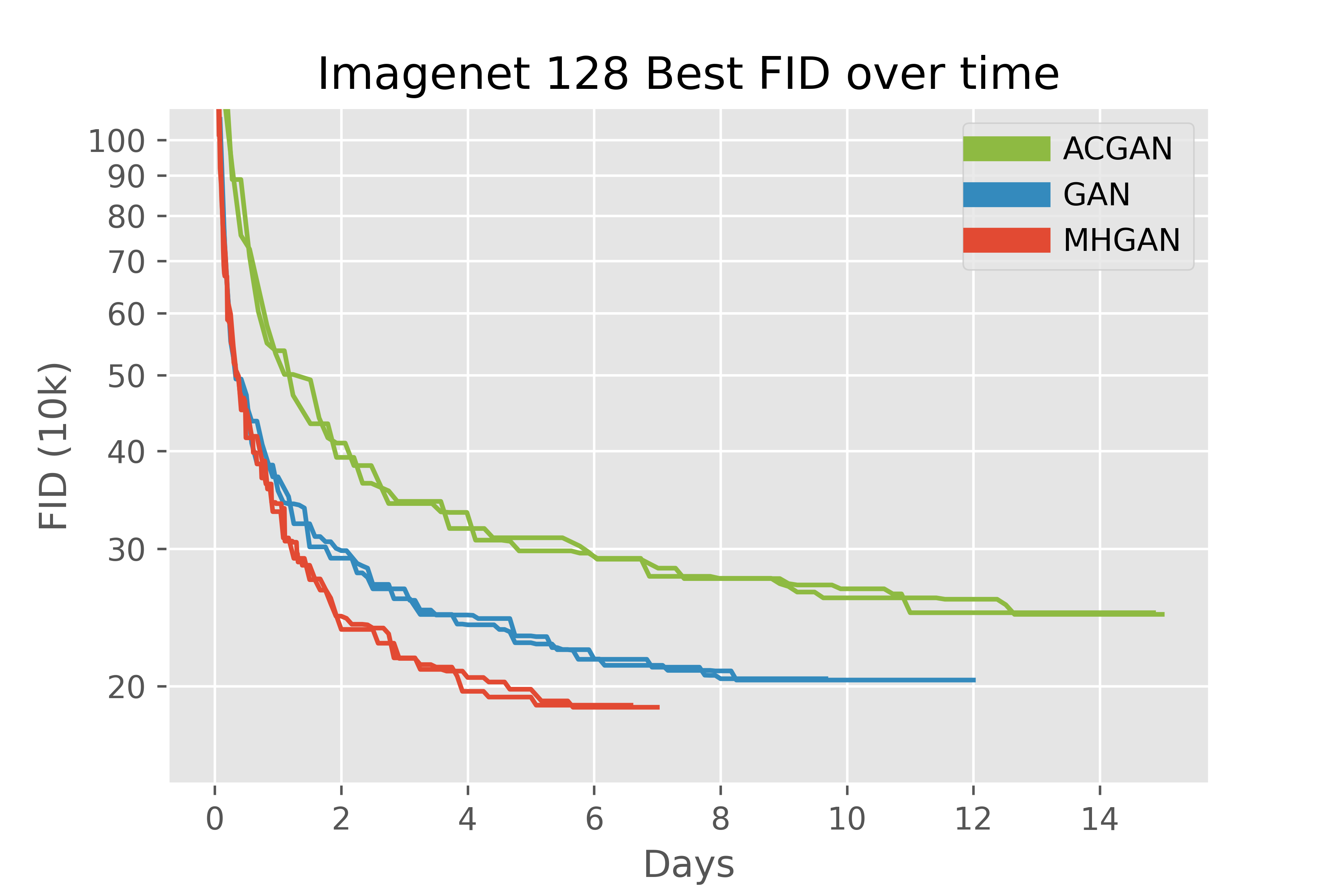}
		\caption{Fully Supervised}
	\end{subfigure}
	\caption{The best FID of our models plotted by duration of training. The multi-hinge loss accelerates training over SAGAN and ACGAN baselines in both fully supervised and semi-supervised settings.}
	\label{fig:bestFIDs}
\end{figure*}

\Cref{fig:bestFIDs} shows that the best FID over time was more quickly achieved by MHGAN-SSL for 1 million iterations, though the performance is not as impressive as in the fully supervised case.

\section{Appendix: Imagenet-64}

\begin{figure*}[htb!]
	\centering
	\begin{subfigure}{.5\textwidth}
		\centering
		\includegraphics[width=1\linewidth]{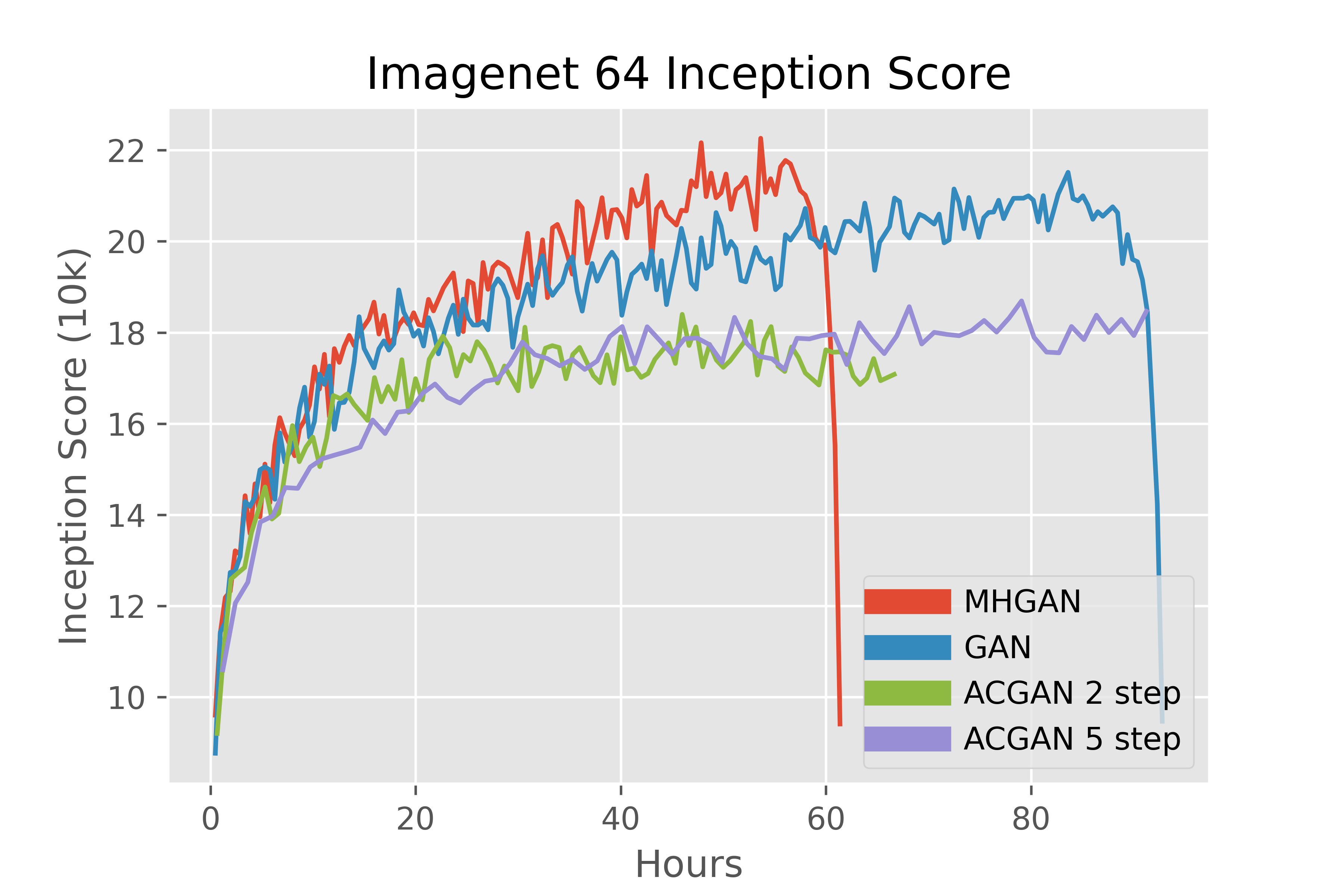}
		\caption{}
		\label{fig:I64_IS}
	\end{subfigure}%
	\hfill
	\begin{subfigure}{.5\textwidth}
		\centering
		\includegraphics[width=1\linewidth]{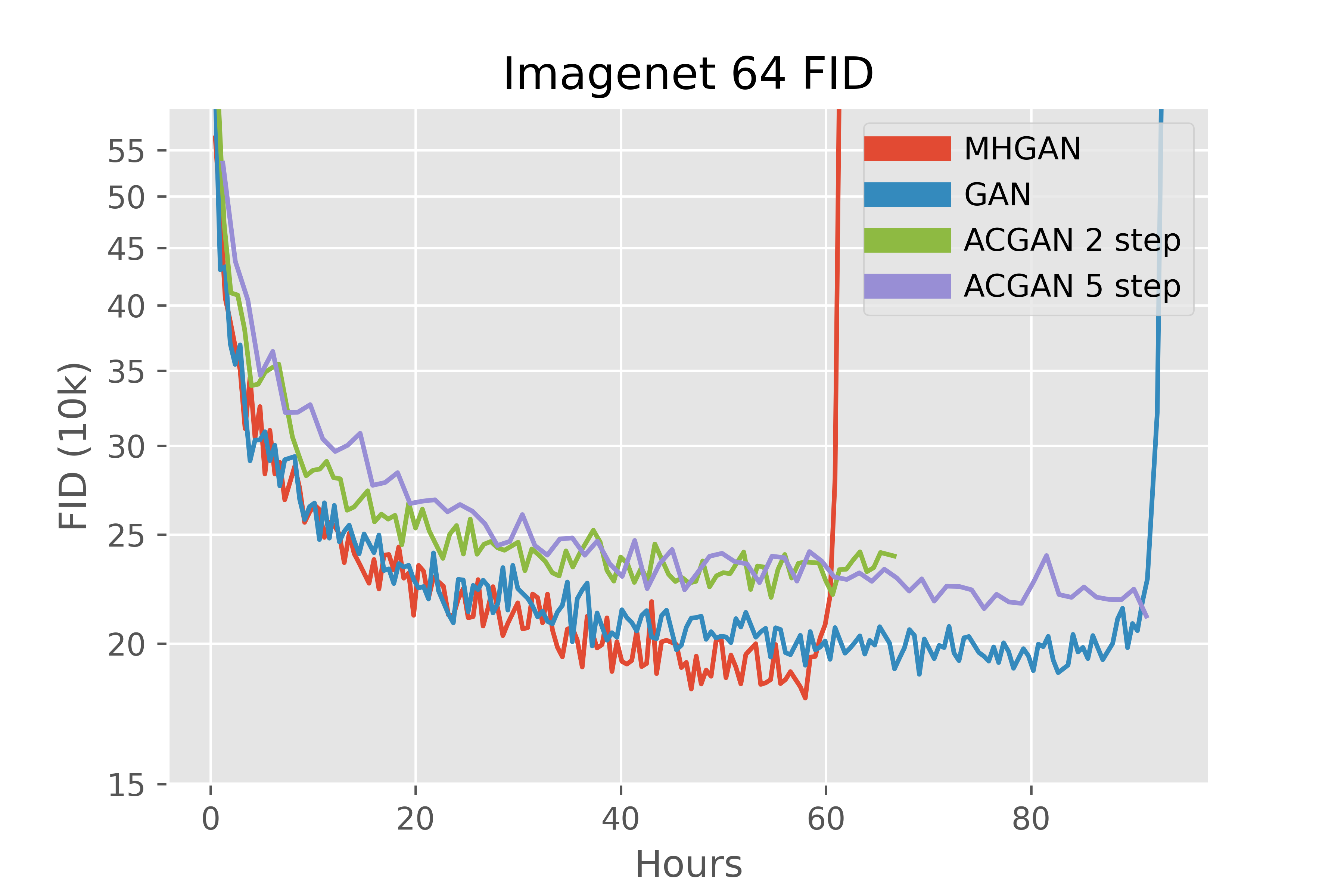}
		\caption{}
		\label{fig:I64_FID}
	\end{subfigure}
	\caption{Inception Scores and FIDs (10k) for supervised image generation on Imagenet $64\times 64$. 1 Million D-steps for SAGAN and MHGAN can be completed in about 48 hours.}
	
\end{figure*}

We also evaluated MHGAN on a smaller scale dataset of Imagenet at resolution $64\times 64$. The conclusions from these experiments are the same but more muted compared to Imagenet at $128\times128$.
\Cref{fig:I64_IS,fig:I64_FID} show that MHGAN improves the FID and IS scores over SAGAN and ACGAN baselines, and in \cref{tab:I64} are FID and IS numbers calculated over 50k images, and the Intra-FID comparison of SAGAN versus MHGAN which shows that diversity stayed the same between the two models.

\begin{table*}[htb!]
	\hfill
	\begin{minipage}{.4\linewidth}
		\centering
		\resizebox{\textwidth}{!}{\begin{tabular}{lcccc}
				\toprule
				\multicolumn{1}{c}{Method} &
				\multicolumn{2}{c}{\textbf{Imagenet 64x64}} \\

				\multicolumn{1}{c}{}    & 
				IS & FID       \\
				\midrule
				Real data  & 56.67 &   \\
				\midrule
				Baseline 1M              & 19.60 & 15.49      \\
				MHGAN 1M           & {\bf 22.16 } & {\bf 13.29}    \\
				\bottomrule
				
		\end{tabular}}
	\end{minipage}%
	\hfill
	\begin{minipage}{.5\linewidth}
		
		\centering
		\includegraphics[width=1\linewidth]{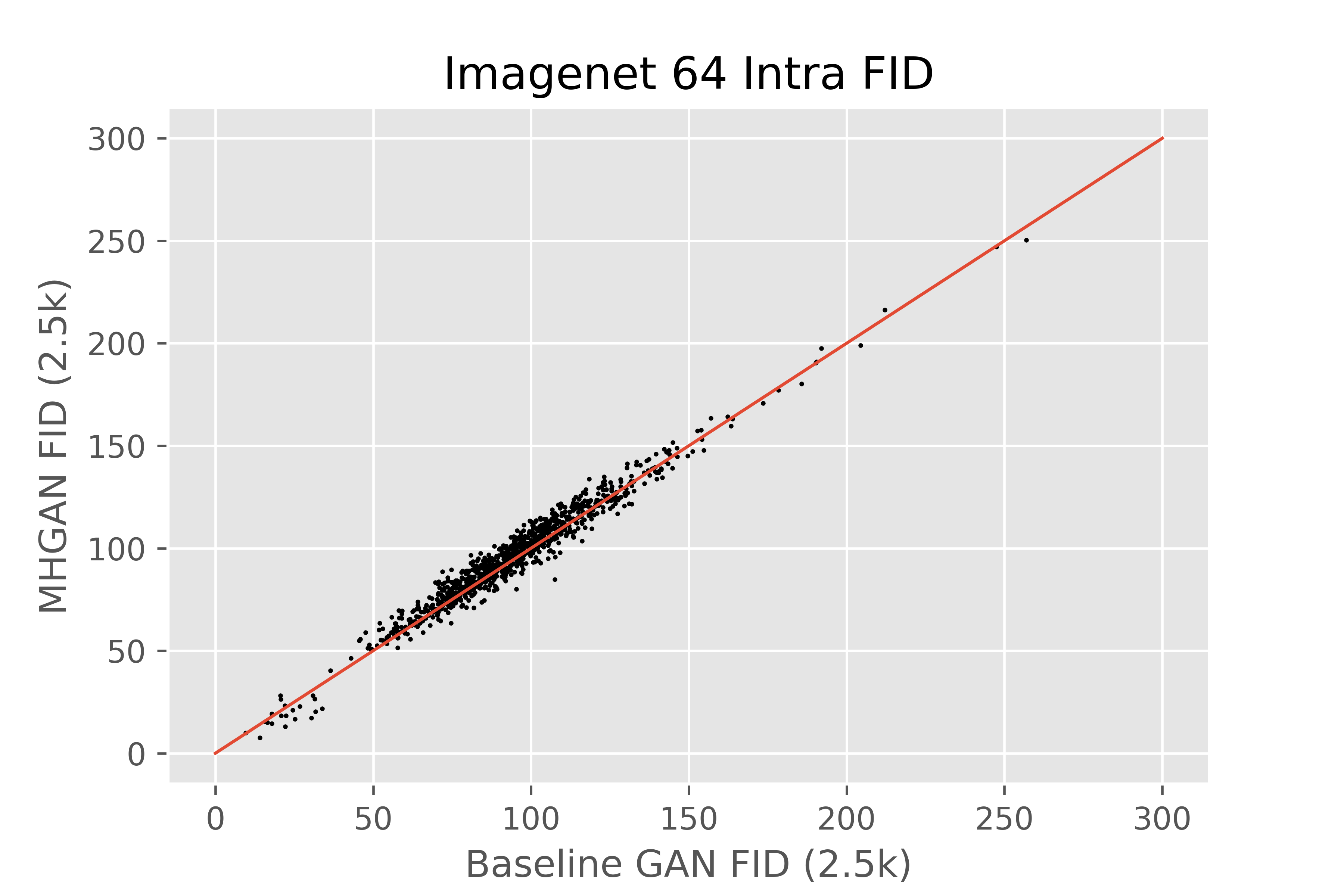}
		
	\end{minipage}
	\caption{Inception Scores and FIDs (50k) for supervised image generation on Imagenet $64\times 64$. Though the margin of improvement is not as great as for Imagenet $128\times128$, adding the auxiliary MHGAN loss proves fruitful. Diversity is maintained at the same level (it improved in Imagenet $128\times128$).}
	\label{tab:I64}
\end{table*}

\subsection{MHShared GAN finetuning for Imagenet $64\times64$}

\begin{figure*}[htb!]
	\centering
	\begin{subfigure}{.33\textwidth}
		\centering
		\includegraphics[width=1\linewidth]{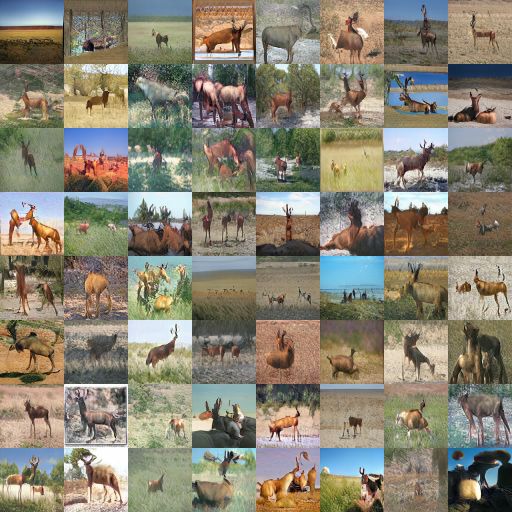}
	\end{subfigure}%
	\hfill
	\begin{subfigure}{.33\textwidth}
		\centering
		\includegraphics[width=1\linewidth]{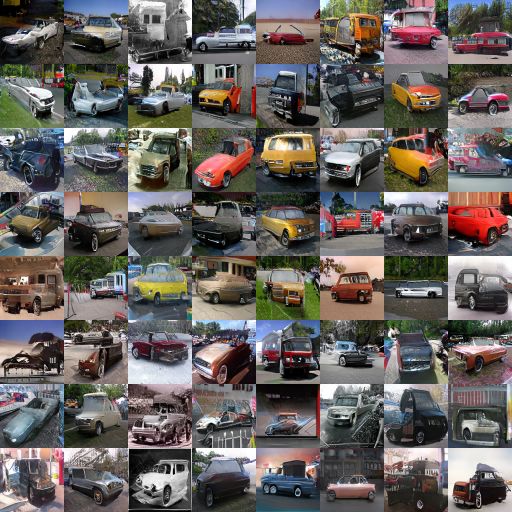}
	\end{subfigure}%
	\hfill
	\begin{subfigure}{.33\textwidth}
		\centering
		\includegraphics[width=1\linewidth]{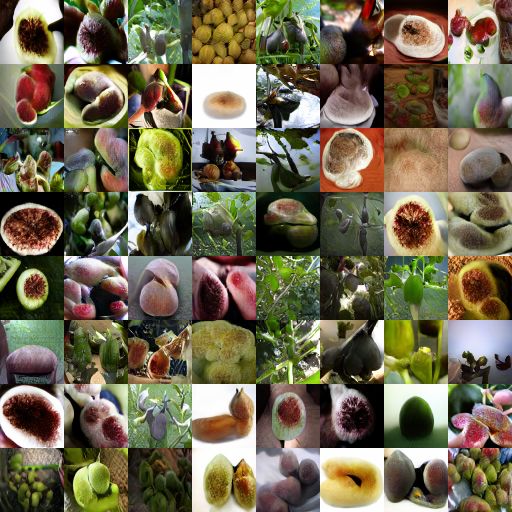}
	\end{subfigure}
	\hfill
	\begin{subfigure}{.33\textwidth}
		\centering
		\includegraphics[width=1\linewidth]{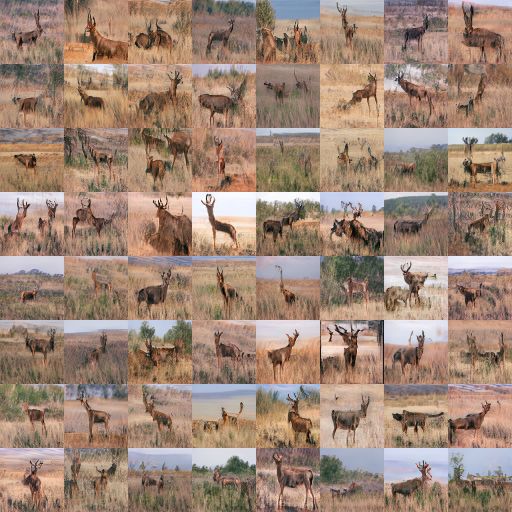}
	\end{subfigure}%
	\hfill
	\begin{subfigure}{.33\textwidth}
		\centering
		\includegraphics[width=1\linewidth]{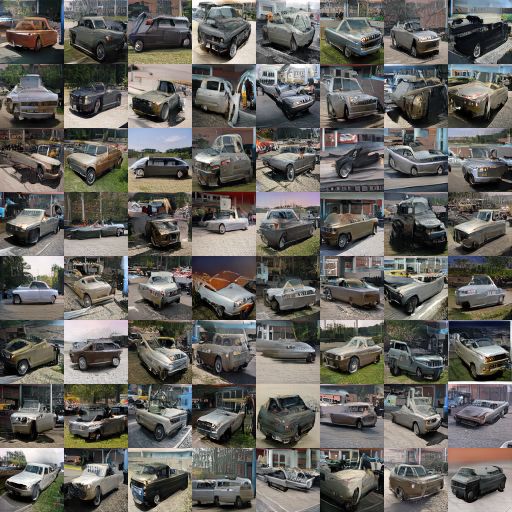}
	\end{subfigure}%
	\hfill
	\begin{subfigure}{.33\textwidth}
		\centering
		\includegraphics[width=1\linewidth]{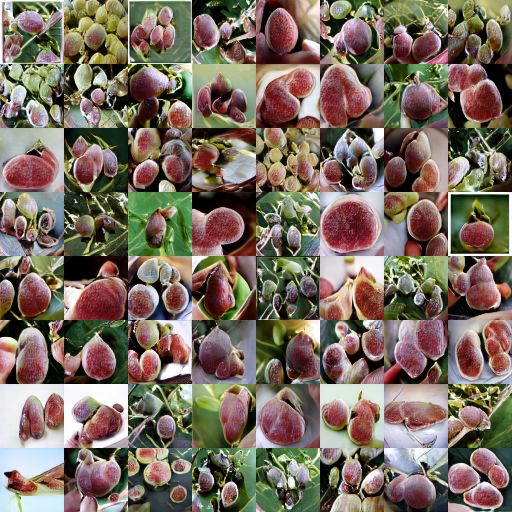}
	\end{subfigure}
	\caption{Before/After performing the MHSharedGAN finetuning to lower the diversity, but raise the quality (according to IS and FID metrics) of samples on Imagenet $64\times64$. The top row of images is sampled from SAGAN trained for 1M steps. And the bottom row is that same model after 15k steps using the MHSharedGAN strategy, where the projection discriminator weights are shared between a itself and a classifier optimized with multi-hinge loss.}
	\label{fig:I64_lowdiv}
\end{figure*}

We also performed the MHShared GAN finetuning on the SAGAN in \Cref{tab:I64} (trained for 1M steps) and after 15k steps the IS went to 30.47 and the FID went to 10.03 (and the validation classification accuracy went to 23.60\% while the generator classification self accuracy went to 62.66\%, both started at $1/n$ class).
Again the diversity of the generator visibly declined; some examples of this are in \Cref{fig:I64_lowdiv}.

\end{document}